\documentclass[11pt,a4paper]{article}
\usepackage[hyperref]{acl2021}
\usepackage{times}
\usepackage{latexsym}
\usepackage{xcolor}
\usepackage{float}
\usepackage{tabularx,booktabs} 
\usepackage{multirow}
\usepackage{xspace}

\usepackage{fancyhdr}
\fancypagestyle{firststyle}
{
   \fancyhf{}
   \fancyfoot[C]{1718 \\ \footnotesize\textit{\newline Proceedings of the 29th International Conference on Computational Linguistics}, pages 1718–1731\\ October 12–17, 2022.}
}
\def\n{1717}
\fancypagestyle{allpage}
{
   \fancyhf{}
   \fancyfoot[C]{\the\numexpr\n+\thepage}
}
\usepackage[disable]{todonotes}
\makeatletter
\newcommand*\iftodonotes{\if@todonotes@disabled\expandafter\@secondoftwo\else\expandafter\@firstoftwo\fi}  % defines \iftodonotes{<true>}{<false>}, thanks to https://tex.stackexchange.com/questions/126559/conditional-based-on-packageoption
\makeatother

\newlength{\extramargin}
\usepackage{calc}
\iftodonotes{\usepackage[paperwidth=\paperwidth+\extramargin*2,marginparwidth=\marginparwidth+\extramargin,width=\textwidth,height=\textheight]{geometry}}{}
\usepackage{microtype}
\usepackage{amsmath,amssymb}
\usepackage{mathtools}

\usepackage{enumitem}

\usepackage{graphicx}
\usepackage{subcaption}

\usepackage{microtype}

\usepackage{siunitx}
\newenvironment{remark}[1][Remark]{\begin{trivlist}
\item[\hskip \labelsep {\bfseries #1}]}{\end{trivlist}}

\newcommand{\nop}[1]{}

\usepackage{amsmath}

\newcommand{\E}{$\mathcal{E\,}$}
\renewcommand{\L}{$\mathcal{L\,}$}
\newcommand{\R}{$\mathcal{R\,}$}
\newcommand{\C}{$\mathcal{C\,}$}

\newcommand{\OurMethod}{ArcaneQA\xspace}

\newcommand{\Freebase}{{\small\textsc{Freebase}}\xspace}
\newcommand{\GoogleKG}{{\small\textsc{Google Knowledge Graph}}\xspace}
\newcommand{\Commons}{{\small\textsc{Commons}}\xspace}
\newcommand{\WebQ}{{\small\textsc{WebQ}}\xspace}
\newcommand{\WebQSP}{{\small\textsc{WebQSP}}\xspace}
\newcommand{\ComplexQ}{{\small\textsc{ComplexWebQ}}\xspace}
\newcommand{\GraphQ}{{\small\textsc{GraphQ}}\xspace}
\newcommand{\GrailQ}{{\small\textsc{GrailQA}}\xspace}

\newenvironment{entityfont}{\fontfamily{qcr}\selectfont\footnotesize}{\par}
\newenvironment{relationfont}{\fontfamily{qcr}\selectfont\footnotesize}{\par}
\DeclareTextFontCommand{\textentity}{\entityfont}
\DeclareTextFontCommand{\textrelation}{\relationfont}
\newcommand\nl[1]{{\it``#1''}} % Natural language

\newcommand{\iid}{i.i.d.\xspace}
\newcommand{\IID}{I.I.D.\xspace}

\aclfinalcopy % Uncomment this line for the final submission
\title{ArcaneQA: Dynamic Program Induction and Contextualized Encoding for Knowledge Base Question Answering}

\author{Yu Gu \\
  The Ohio State University \\
  Columbus, Ohio, USA \\
  \texttt{gu.826@osu.edu} \\\And
  Yu Su\\
  The Ohio State University \\
  Columbus, Ohio, USA \\
  \texttt{su.809@osu.edu} \\}

\date{}

\begin{document}
\maketitle
\thispagestyle{firststyle}
\begin{abstract}

\nop{Question answering on knowledge bases (KBQA) provides a user-friendly interface to the massive structured knowledge in KBs. It presents an array of intertwined challenges including non-\iid\ generalization, large search space, and schema linking. Existing studies have tackled some of these challenges, but a unified solution that can address all the challenges is still lacking, which limits the applicability of KBQA on large-scale KBs. In this work, we propose a novel model, \OurMethod, that aims to tackle all the aforementioned challenges and achieve strong (non-\iid) generalization on large-scale KBs. Our model is based on two mutually-boosting components: \textit{dynamic program induction}, which dynamically prunes the search space, and \textit{dynamic contextualized encoding}, which guides the search process on the large-scale KB via implicit schema linking. Experiment results on multiple popular KBQA datasets demonstrate the highly competitive performance of \OurMethod in both effectiveness and efficiency, especially on non-\iid\ generalization and complex questions involving multiple relations and functions.\footnote{We will release our code upon acceptance.}}
Question answering on knowledge bases (KBQA) poses a unique challenge for semantic parsing research due to two intertwined challenges: \textit{large search space} and \textit{ambiguities in schema linking}. Conventional ranking-based KBQA models, which rely on a candidate enumeration step to reduce the search space, struggle with flexibility in predicting complicated queries and have impractical running time. In this paper, we present \OurMethod, a novel generation-based model that addresses both the large search space and the schema linking challenges in a unified framework with two mutually boosting ingredients: \textit{dynamic program induction} for tackling the large search space and \textit{dynamic contextualized encoding} for schema linking. Experimental results on multiple popular KBQA datasets demonstrate the highly competitive performance of \OurMethod in both effectiveness and efficiency.\footnote{Data and code: \href{https://github.com/dki-lab/ArcaneQA}{dki-lab/ArcaneQA}} 

\end{abstract}

\section{Introduction}

Modern knowledge bases (KBs) contain a wealth of structured knowledge. For example, \Freebase~\cite{bollacker2008freebase} contains over \num{45} million entities and \num{3} billion facts across more than \num{100} domains, while \GoogleKG has amassed over \num{500} billion facts about \num{5} billion entities~\cite{googlekg2020}. Question answering on knowledge bases (KBQA) has emerged as a user-friendly solution to access the massive structured knowledge in KBs. 

\begin{figure}[!th]
\centering
\includegraphics[width=1.0\linewidth]{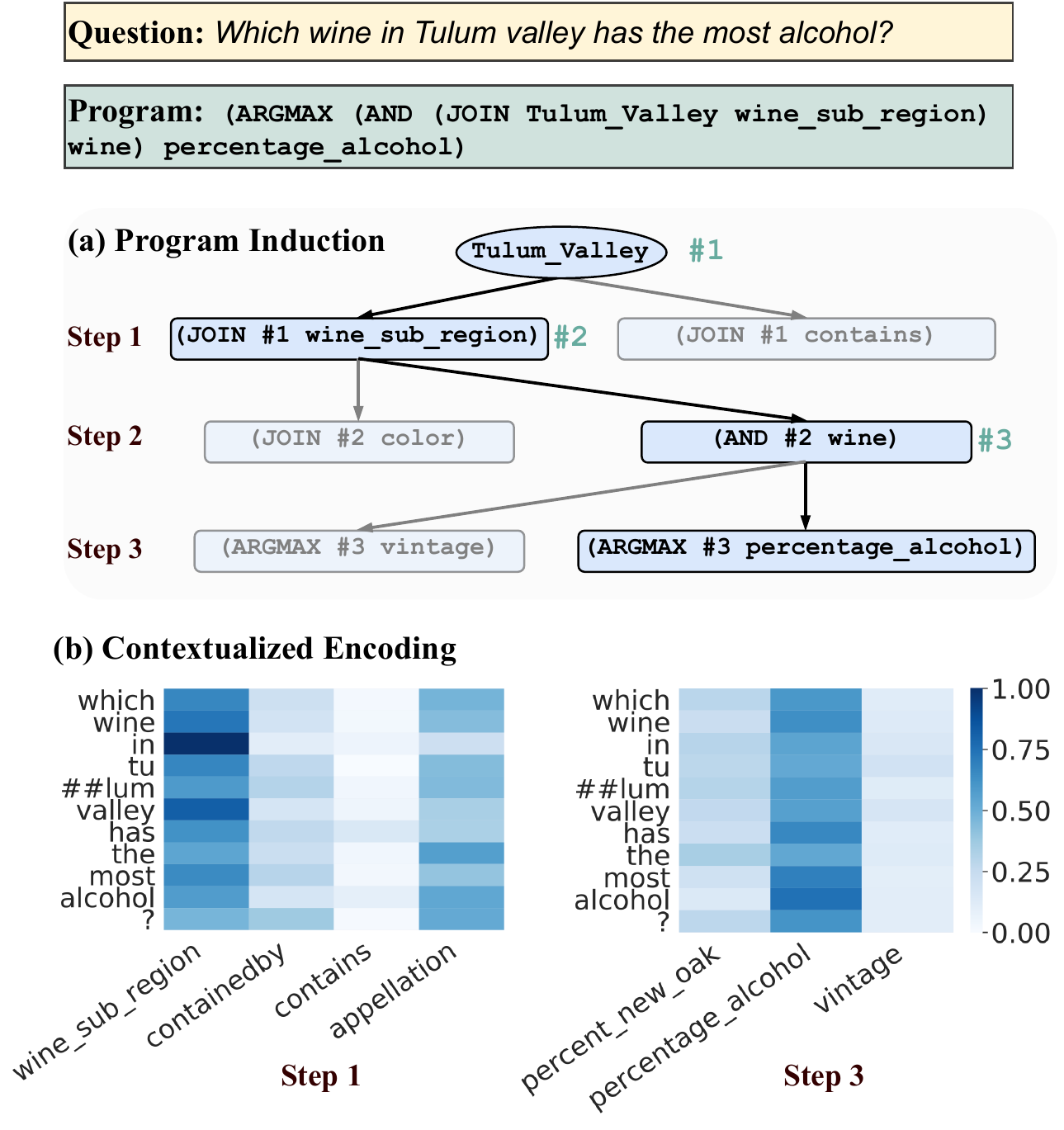}
% \caption{A high-level illustration of our program induction procedure. We predict the target program by incrementally synthesizing a sequence of subprograms (denoted as rounded rectangles). Every time we predict a subprogram, we execute it and denote it as a variable (denoted as eclipses, \textrelation{V1-3}). The initial variable \textrelation{V1} is an entity mentioned in the question. A variable further guides the generation of subsequent programs. Solid arrows mean synthesizing a new (sub)program from a variable, while dotted arrows mean executing a (sub)program into a variable. 
% }
\caption{KBQA is commonly modeled as semantic parsing with the goal of mapping a question into an executable program. \textbf{(a)} A high-level illustration of our program induction procedure. The target program is induced by incrementally synthesizing a sequence of subprograms (\textrelation{\#1-3}). The execution of each subprogram can significantly reduce the search space of subsequent subprograms. \textbf{(b)} Alignments between question words and schema items at different steps achieved by a BERT encoder. A pre-trained language model like BERT can jointly encode the question and schema items to get the contextualized representation at each step, which further guides the search process.}
\label{fig:intro}
% \vspace{-10pt}
\end{figure}

KBQA is commonly modeled as a semantic parsing problem~\cite{zelle1996learning,zettlemoyer2005learning} with the goal of mapping a natural language question into a logical form that can be executed against the KB~\cite{berant-etal-2013-semantic,cai-yates-2013-semantic,yih-etal-2015-semantic}. Compared with other semantic parsing settings such as text-to-SQL parsing~\cite{wikisql, yu-etal-2018-spider}, where the underlying data is moderate-sized, the massive scale and \nop{the semantic ambiguities arising from the broad schema} the broad-coverage schema of KBs makes KBQA a uniquely challenging setting for semantic parsing research.

\nop{However, most early studies on KBQA still (implicitly) operate with the conventional \emph{\iid}\ assumption, i.e., assuming test questions are drawn from the same distribution as the training data. \citet{gu2020iid} have recently pointed out that the \iid\ assumption may not be practical for large-scale KBs due to the enormous space and KBQA models should be built with generalizability to novel compositions of scheme items (\emph{compositional generalization}) and unseen schema items (\emph{zero-shot generalization}).

Developing KBQA models with strong generalizability on large-scale KBs is a highly non-trivial task.}

The unique difficulty stems from two intertwined challenges: \textit{large search space} and \emph{ambiguities in schema linking}. On the one hand, transductive semantic parsing models that are highly effective in other semantic parsing settings~\cite{dong2016language,wang-etal-2020-rat} struggle with the large vocabulary size and often generate logical forms (i.e., formal queries)\footnote{We use the terms logical form, query, and program interchangeably across the paper.} that are not faithful to the underlying KB~\cite{gu2020iid,xie2022unifiedskg}. Therefore, a candidate enumeration and ranking approach is commonly adopted for KBQA~\cite{berant2014semantic, yih-etal-2015-semantic,Abujabal, lan2019knowledge, sun2020sparqa, gu2020iid, ye2021rng}.
% where a set of candidate logical forms are first generated from the KB and are then ranked by a semantic matching model
However, these methods have to make various compromises on the complexity of admissible logical forms to deal with the large search space. Not only does this limit the type of answerable questions, but it also leads to impractical runtime performance due to the time-consuming candidate enumeration~\cite{gu2020iid}. 
% \nop{These issues already present themselves under the \iid\ setting and are further amplified when a model needs to generalize in a non-\iid\ fashion.} 
On the other hand, schema linking,\footnote{Semantic parsing implicitly entails two sub-tasks: \textit{schema linking} and \textit{composition}. There is not necessarily a dedicated step or component for schema linking. More commonly, the two sub-tasks are handled simultaneously.} i.e., mapping natural language to the corresponding schema items in the KB (e.g., in \autoref{fig:intro}, \textrelation{wine\_sub\_region} is a linked schema items), is also a core challenge of KBQA. Compared with text-to-SQL parsing~\cite{hwang2019comprehensive, zhang2019editing, wang-etal-2020-rat}, the broad schema of KBs and the resulting ambiguity between schema items makes accurate schema linking more important and challenging for KBQA. Recent studies show that contextualized joint encoding of natural language questions and schema items with BERT~\cite{devlin-etal-2019-bert} can significantly boost the schema linking accuracy~\cite{gu2020iid,chen2021retrack}. 
% \nop{\citet{gu2020iid} recently show that contextualized joint encoding of natural language question and schema items with BERT~\cite{devlin-etal-2019-bert} significantly improves compositional and zero-shot generalization.} 
However, existing methods still struggle with the large search space and need to encode a large number of schema items, which is detrimental to both accuracy and efficiency.

\nop{TODO: I did't change the motivation much (i.e., addressing the above two challenges), because the analyses of these two challenges properly motivate the high-level design of our model. Also, when introducing our model, I will still make comparisons with other generation-based models and briefly discuss the advantages of our model, but I won't claim that previous generation-based models somewhat have some fundamental problems and motivate our model in that way, which is weak and may draw attacks.}

We present \OurMethod (Dyn\underline{\textbf{a}}mic P\underline{\textbf{r}}ogram Indu\underline{\textbf{c}}tion \underline{\textbf{a}}nd Co\underline{\textbf{n}}textualized \underline{\textbf{E}}ncoding for \underline{\textbf{Q}}uestion \underline{\textbf{A}}nswering), a \textit{generation-based} KBQA model that addresses both the large search space and the schema linking challenges in a unified framework. Compared with the predominant ranking-based KBQA models, our generation-based model can prune the search space on the fly and thus is more flexible to generate diverse queries without compromising the expressivity or complexity of answerable questions. Inspired by prior work~\cite{dong2016language, liang-etal-2017-neural,SMDataflow2020,chen2021retrack}, we model KBQA using the encoder-decoder framework.
% which is the de facto architecture in many other semantic parsing tasks. 
However, instead of top-down decoding with grammar-level constraints as in prior work,
% ~\cite{dong2016language,chen2021retrack}, 
which does not guarantee the faithfulness of the generated queries to the underlying KB, \OurMethod performs \textit{dynamic program induction}~\cite{liang-etal-2017-neural, SMDataflow2020}, where we incrementally synthesize a program by dynamically predicting a sequence of subprograms to answer a question; i.e., \textit{bottom-up parsing}~\cite{cheng2019learning,rubin-berant-2021-smbop}. Each subprogram is grounded to the KB and its grounding (i.e., denotation or execution results) can further guide an efficient search for faithful programs (see \autoref{fig:intro}(a)). In addition, we unify the meaning representation (MR) for programs in KBQA using S-expressions and support more diverse operations over the KB (e.g., numerical operations such as \textrelation{COUNT/ARGMIN/ARGMAX} and diverse graph traversal operations).\nop{we unify the meaning representation (MR) for queries in KBQA and support more diverse operations than prior work~\cite{liang-etal-2017-neural, lan-jiang-2020-query}.} 

At the same time, we employ pre-trained language models (PLMs) like BERT to jointly encode the question and schema items and get the contextualized representation of both, which implicitly links words to the corresponding schema items via self-attention. One unique feature to note is that \textit{the encoding is also dynamic}: at each prediction step, only the set of admissible schema items determined by the dynamic program induction process needs to be encoded, which allows extracting the most relevant information from the question for each prediction step while avoiding the need to encode a large number of schema items. ~\autoref{fig:intro}(b) illustrates the contextualization of different steps via the attention heatmaps of BERT. In this example, the attention of each question word over candidate schema items serves as a strong indicator of the gold items for both steps (i.e., \textrelation{wine\_sub\_region} for step 1 and \textrelation{percentage\_alcohol} for step 3). The two key ingredients of our model are \textit{mutually boosting}: dynamic program induction significantly reduces the number of schema items that need to be encoded, while dynamic contextualized encoding intelligently guides the search process.

Our main contribution is as follows: a) We propose a novel generation-based KBQA model that is flexible to generate diverse complex queries while also being more efficient than ranking-based models. b) We propose a novel strategy to effectively employ PLMs to provide contextualized encoding for KBQA. c) We unify the meaning representation (MR) of different KBQA datasets and support more diverse operations. d) With our unified MR, we evaluate our model on three popular KBQA datasets and show highly competitive results.

\nop{We evaluate our model on a number of popular KBQA datasets. On datasets that require non-\iid\ generalization, including \GrailQ~\cite{gu2020iid} and \GraphQ~\cite{su-etal-2016-generating}, \OurMethod achieves new state-of-the-art performance, outperforming the prior art by 3\% to 6.9\% absolute in F1. On the \WebQ~\cite{berant-etal-2013-semantic} dataset that mainly focuses on \iid\ generalization, our model also performs competitively. Fine-grained analyses further demonstrate \OurMethod's advantages in handling complex questions with multiple relations and functions as well as runtime efficiency.}

\section{Related Work}
\begin{remark} [Ranking-Based KBQA.]
To handle the large search space in KBQA, existing studies typically rely on hand-crafted templates with a pre-specified maximum number of relations to enumerate candidate logical forms~\cite{yih-etal-2015-semantic, Abujabal, lan2019knowledge, bhutani2019learning, bhutani-etal-2020-answering}, which suffers from limited expressivity and scalability. For example, \citet{yih-etal-2015-semantic} limit the candidate programs to be a core relational chain, whose length is at most two, plus constraints. \citet{ye2021rng} additionally adopts a post-generation module to revise the enumerated logical forms into more complicated ones, however, their method still heavily depends on the candidate enumeration step. In addition, the time-consuming candidate enumeration results in impractical online inference time for ranking-based models. In contrast, \OurMethod obviates the need for candidate enumeration by pruning the search space on the fly and thus can generate more diverse and complicated programs within practical running time.
\end{remark}

\begin{remark} [Generation-Based KBQA.]
To relax the restriction on candidate enumeration, some recent efforts are made to reduce the search space using beam search~\cite{lan2019multi, chen2019uhop, lan-jiang-2020-query}, however, \citet{lan2019multi} and \citet{chen2019uhop} can only generate programs of path structure, while \citet{lan-jiang-2020-query} follow the query graph structure proposed by~\citet{yih-etal-2015-semantic}. A few recent studies~\cite{liang-etal-2017-neural,chen2021retrack} formulate semantic parsing over the KB as sequence transduction using encoder-decoder models to enable more flexible generation. \citet{chen2021retrack} apply schema-level constraints to eliminate ill-formed programs from the search space, however, they do not guarantee the faithfulness of predicted programs. Similar to \citet{liang-etal-2017-neural}, our dynamic program induction uses KB contents-level constraints to ensure the faithfulness of generated programs, but we extend it to handle more complex and diverse questions and also use it jointly with dynamic contextualized encoding.
\end{remark}

\begin{remark}[Using PLMs in Semantic Parsing.]
PLMs have been widely applied in many semantic parsing tasks, typically being used to jointly encode the input question and schema items\cite{hwang2019comprehensive, zhang2019editing, wang-etal-2020-rat,scholak2021picard}. However, PLMs have been under-exploited in KBQA. One major difficulty of using PLMs in KBQA lies in the high volume of schema items in a KB; simply concatenating all schema items with the input question for joint encoding, as commonly done in text-to-SQL parsing, will vastly exceed PLMs' maximum input length. Existing KBQA models either use PLMs to provide features for downstream classifiers\cite{lan-jiang-2020-query,sun2020sparqa} or adopts a pipeline design to identify a smaller set of schema items beforehand and only use PLMs to encode these identified items~\cite{gu2020iid,chen2021retrack}, which can lead to error propagation. By comparison, \OurMethod can fully exploit PLMs to provide contextualized representation for the question and schema items dynamically, where only the most relevant schema items are encoded at each step. More recently, \citet{ye2021rng} use T5~\cite{raffel2019exploring} to output a new program given a program as input, while T5's decoder generates free-formed text and does not always produce faithful programs. By contrast, \OurMethod only uses PLMs for encoding and uses its customized decoder with a faithfulness guarantee.
\end{remark}

\nop{\begin{remark} [Other Related Work.]
Other related works include grammar-constrained semantic parsing~\cite{krishnamurthy-etal-2017-neural, yin-neubig-2017-syntactic} that relies on type-constrained grammar to generate well-formed programs. \OurMethod also uses grammar rules to guarantee well-formedness but additionally uses fact-level information to guarantee the faithfulness of the predicted programs. There is also an interesting connection to learning to search~\cite{daume2005learning} because \OurMethod learns to dynamically interact with a large environment to generate a structured output.
\end{remark}}

\section{Background}
\begin{remark}[Knowledge Base.]
A knowledge base $\mathcal{K}$ consists of a set of relational triplets $\mathcal{K}_r\subset\mathcal{E}\times\mathcal{R}\times(\mathcal{E}\cup\mathcal{L})$ and a set of class assertions $\mathcal{K}_c\subset\mathcal{E}\times\mathcal{C}$, where \C is a set of classes, \E is a set of entities, \L is a set of literals and \R is a set of binary relations. Elements in \C and \R are also called the schema items of $\mathcal{K}$.
% Based on the definition, triplets in $\mathcal{K}$ are directional. For simplicity, we add new triplets to the KB by reversing the original relation and adding a prefix ``R\_" to make every triplet bidirectional, so we can ignore the directions of edges in the following of this paper.
\end{remark}

\begin{remark}[Meaning Representation for KBQA.]
Prior work adopt different meaning representations to represent logical forms for KBQA. For example, \citet{yih-etal-2015-semantic} use graph query, which represents a program as a core relation chain with (optionally) some entity constraints. \citet{cai-yates-2013-semantic} use $\lambda$-Calculus as their meaning representation. In this paper, we follow~\citet{gu2020iid} to use S-expressions as our meaning representation due to their expressivity and simplicity. To support more diverse operations over the KB, we extend their definitions with two additional functions \textrelation{CONS} and \textrelation{TC}, which are used to support constraints with implicit entities and temporal constraints respectively (see details in \autoref{appendix:mr}). For implicit entities, consider the question \nl{What was Elie Wiesel's father's name?}, whose target query involves two entities: \textentity{Elie\_Wiesel} and \textentity{Male}. The entity \textentity{Male} is an implicit constraint rather than a named entity,\footnote{\WebQSP is the only dataset we consider that has this feature. Though there might be a more systematic way to differentiate implicit entities from named entities, we choose an expedient way to collect implicit entities from the training data according to whether an entity is explicitly mentioned in the question.} and it is used as an argument of \textentity{CONS} in the target logical form: \textrelation{(CONS (JOIN people.person.children Elie\_Wiesel) people.person.gender Male)}. \textentity{TC} works in a similar way, with the difference being that the constraint should be a temporal expression (e.g., 2015-08-10) rather than an implicit entity. \nop{Our definitions can almost support all kinds of queries in existing KBQA datasets. We further provide existing datasets with annotations in our meaning representation to benefit future research.\footnote{Our annotations for \WebQSP\cite{yih-etal-2016-value} have already been used in~\citet{chen2021retrack,ye2021rng}}}
\end{remark}

\nop{\begin{remark}[KBQA as Program Induction.]
We treat KBQA as program induction~\cite{liang-etal-2017-neural,SMDataflow2020}, in which we seek to map a question onto a program to retrieve answers from the KB. We choose the S-expression formulation used in \citet{gu2020iid}, which can be easily decomposed into a sequence of expressions by de-nesting. There are five types of functions defined in S-expression, including \textrelation{COUNT} that returns the cardinality of a set, \textrelation{AND} that denotes set intersection, \textrelation{JOIN} that extends from one set of entities to another set via a relation,
% \footnote{It typically means walking from a set of entities to another set of entities via a certain relation} 
superlatives (\textrelation{ARGMAX/ARGMIN}), and comparatives (\textrelation{LT/LE/GT/GE}). For more details, we refer the reader to~\citet{gu2020iid}. For the other datasets in our experiments, we convert the original representation into S-expression and will release the converted data for future research (Appendix~\ref{sec:anno}).

% The program that corresponds to a complex question can be constructed iteratively from a line of sub-programs. Table~\ref{tab:query_exp} shows an example of a complex question in \GrailQ. To generate the target program, we can first generate sub-program 1 from the identified entity \textentity{C\#}. Next, we formulate sub-program 2 by adding a type constraint to the execution of sub-program 1. Finally, we apply an aggregation operation of argmax to the execution of sub-program 2 and generate the target program. This type of program is referred as S-expression in \GrailQ. We follow the definition of S-expression in \GrailQ and extend it by introducing two new operations \textrelation{TC} and \textrelation{CONS} to generalize our program representation to more datasets. A comprehensive definition is in Table~\ref{tab:query_def}.  We also provide the S-expression program annotations for datasets without such annotation to benefit future research.  More details on the definition of S-expression and annoatation process can be found in Appendix~\ref{}.
\end{remark}}
\section{Approach}
\begin{figure*}[th]
\centering
\includegraphics[width=1\linewidth]{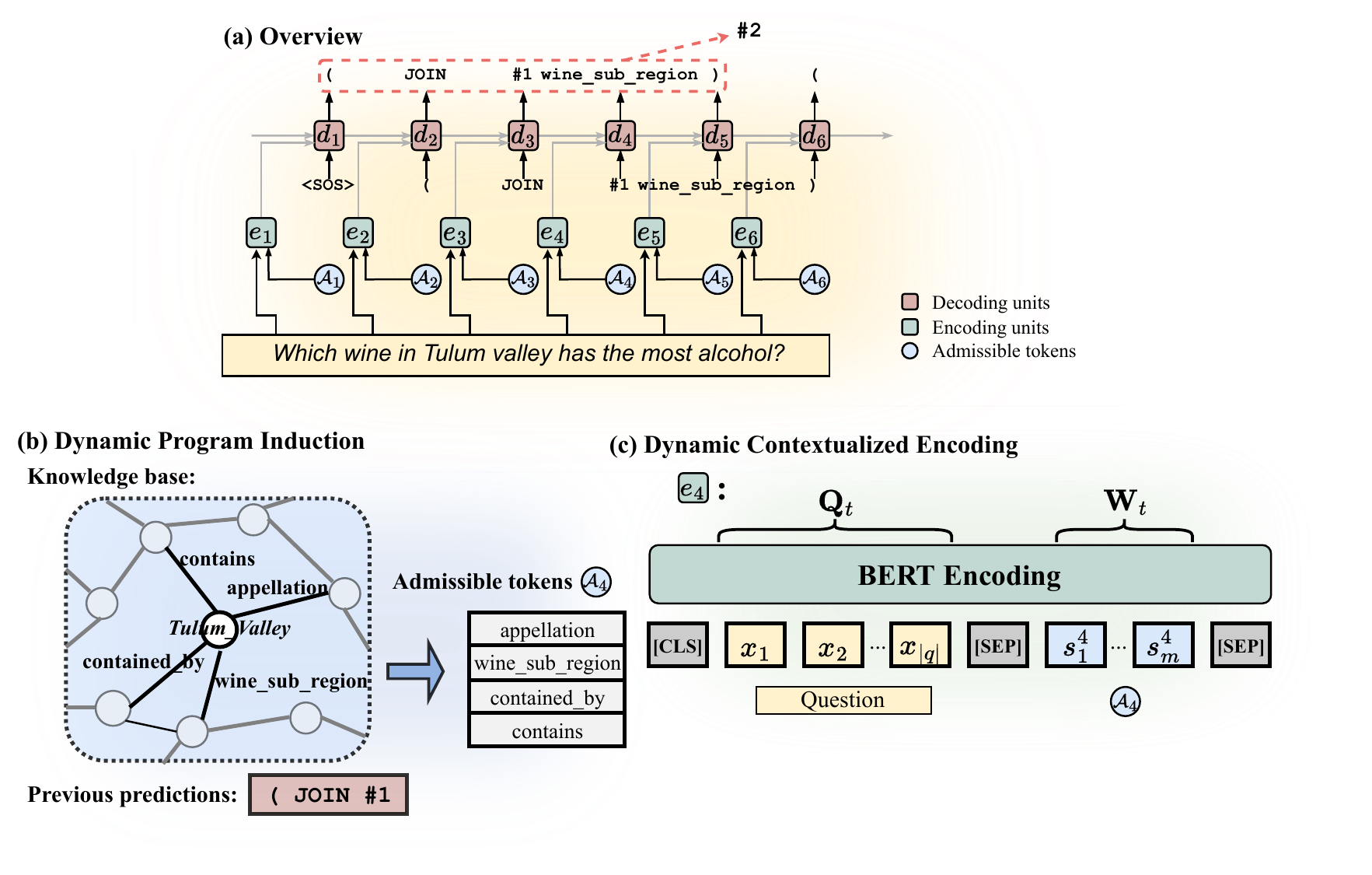}
\caption{\textbf{(a)} Overview of \OurMethod. \OurMethod synthesizes the target program by iteratively predicting a sequence of subprograms.  \textbf{(b)} At each step, it makes a prediction from a small set of admissible tokens $\mathcal{A}$ dynamically determined based on the execution of previous subprograms (for faithfulness to the KB) as well as the grammar (for well-formedness). \textbf{(c)} \OurMethod also leverages BERT to provide dynamic contextualized encoding of the question and the admissible tokens at each step, which enables implicit schema linking and guides the dynamic program induction process.}
\label{fig:overview}
% \vspace{-10pt}
\end{figure*}
% \begin{figure}[th]
% \centering
% \includegraphics[width=\linewidth]{acl-ijcnlp2021-templates/figs/encoding.pdf}
% \caption{}
% \label{fig:encoding}
% \vspace{-10pt}
% \end{figure}
\subsection{Overview}
The core idea of our generation-based model is to gradually expand a subprogram (i.e., a partial query) into the finalized target program, instead of enumerating all possible finalized programs from the KB directly, which suffers from combinatorial explosion. There are two common strategies to instantiate the idea of gradual subprogram expansion, depending on the type of meaning representation being used. For a graph-like meaning representation, we can directly perform graph search over the KB to expand a subprogram~\cite{chen2019uhop,lan-jiang-2020-query}. Also, we can linearize a program into a sequence of tokens and perform decoding in the token space~\cite{liang-etal-2017-neural,scholak2021picard}. Because S-expressions can be easily converted into sequences of tokens, we choose to follow the second strategy and take advantage of the encoder-decoder framework, which has been a de facto choice for many semantic parsing tasks. Concretely, \OurMethod learns to synthesize the target program by dynamically generating a sequence of subprograms token by token until predicting $\langle EOS\rangle$, where each subsequent subprogram is an expansion from one or more preceding subprograms (denoted as parameters in the subsequent subprogram). Formally, the goal is to map an input question $q = x_1, ..., x_{|q|}$ to a sequence of subprograms $o = o^1_1, ..., o^1_{|o^1|}, ..., o^k_1, ..., o^k_{|o^k|} = y_1, ..., y_{|o|}$,
% \ysu{it'd be probably more clear to use superscript for expression id and subscript for token id within expression, like $o^1_1$} 
where $k$ is the number of total subprograms and $|o| = \sum\limits_{i=1}^k|o^i|$. 
% For example, to predict the target program in Figure~\ref{fig:intro},  \OurMethod will predict the following sequences of tokens: \textit{``(", ``JOIN", ``V1", ``wine\_sub\_region", ``)", ``(", ``AND", ``V2", ``wine", ``)", ``(", ``ARGMAX", ``V3", ``percentage\_alcohol", ``)", ``$\langle EOS\rangle$"}, where $\langle EOS\rangle$ denotes the end of decoding.
We base \OurMethod on Seq2Seq with attention~\cite{Seq2Seq, DBLP:journals/corr/BahdanauCB14}, in which the conditional probability $p(o|q)$ is decomposed as:
\begin{equation}
    p(o|q) = \prod_{t=1}^{|o|}p(y_t|y_{<t}, q),
\end{equation}
where each token $y_t$ is either a token from the vocabulary $\mathcal{V}$  or an intermediate subprogram from the set $\mathcal{S}$ storing all previously generated subprograms. $\mathcal{V}$ comprises all schema items in $\mathcal{K}$, syntactic symbols in S-expressions (i.e., parentheses and function names), and the special token $\langle EOS\rangle$. $\mathcal{S}$ initially contains the identified entities in the question (e.g., \textrelation{\#1} in \autoref{fig:intro}). Every time a subprogram is predicted\nop{(i.e., when the model outputs ``)'')}, it is executed and added to $\mathcal{S}$ (e.g., \textrelation{\#2} in \autoref{fig:intro}).

\OurMethod builds on two key ideas: \textit{dynamic program induction} and \textit{dynamic contextualized encoding} (see \autoref{fig:overview}). At each decoding step, \OurMethod only makes a prediction from a small set of admissible tokens instead of the entire vocabulary. This is achieved by the dynamic program induction framework (\autoref{sec:execution}), which effectively prunes the search space by orders of magnitude and guarantees that the predicted programs are faithful to the KB. In addition, we dynamically apply BERT to provide contextualized joint encoding for both the question and admissible tokens at each decoding step (\autoref{sec:plm}). In this way, we allow the contextualized encoding to only focus on the most relevant information without introducing noise from irrelevant tokens.

% Specifically, we rely on intermediate variables to reduce the search space by orders of magnitude, and in this way, we guarantee that the generated program can be grounded to the knowledge base (Section~\ref{sec:execution}). Moreover, we propose a new way to use PLM to dynamically encode the question and schema items at every decoding step (Section~\ref{sec:plm}). Finally, we incorporate the same search space pruning strategy during training and train \OurMethod with strong supervision (Section~\ref{sec:train}). \ygu{TODO: make it clear there are three types of admissible tokens}

\subsection{Dynamic Program Induction} 
\label{sec:execution}
Dynamic program induction capitalizes on the idea that a complicated program can be gradually expanded from a list of subprograms.  To ensure the expanded program is faithful to the KB, we query the KB with a subprogram to expand and a function defined in \autoref{tab:mr} to get a set of admissible actions (tokens). For example, in \autoref{fig:overview}, given \textrelation{\#1} and the function \textrelation{JOIN}, the admissible actions are defined by predicting a relation connecting to the execution result of \textrelation{\#1} (i.e., \textentity{Tulum\_Valley}), and there are only four relations to choose from (e.g., \textrelation{appellation} and \textrelation{wine\_sub\_region}). \autoref{tab:cons} shows a comprehensive description of expansion rules for different functions. With these rules, \OurMethod can greatly reduce the search space for semantic parsing over the KB dynamically. The reduced candidate space further allows us to perform dynamic contextualized encoding (\autoref{sec:plm}).

Within our encoder-decoder framework, this idea is implemented using constrained decoding~\cite{liang-etal-2017-neural,scholak2021picard}, i.e., at each decoding step, a small set of admissible tokens from the vocabulary is determined based on the decoding history following predefined rules. The expansion rules in \autoref{tab:cons} have already comprised part of our rules for constrained decoding. In addition, several straightforward grammar rules are applied to ensure that the generated programs are well-formed. For instance, after predicting ``(", the admissible tokens for the next step can only be a function name. After predicting a function name, the decoder can only choose a preceding subprogram to expand. After predicting ``)", the admissible tokens for next step can only be either ``(", indicating the start of a new subprogram, or ``$\langle EOS\rangle$", denoting termination. The decoding process can be viewed as a \textit{sequential decision-making process}, which decomposes the task of finding a program from the enormous search space into making decisions from a sequence of smaller search spaces.

\begin{table*}[t]
    \centering
    \small
    \resizebox{0.75\linewidth}{!}{\begin{tabular}{cl}
    \toprule
     \textbf{Current function}  & \textbf{Admissible actions} \\
     \midrule
     \textrelation{JOIN} & $\{r|h\in\textrelation{\#}, (h, r, t)\in\mathcal{K}_r\}$\\
     \textrelation{AND} & $\{v|v\in\mathcal{S}, v\cap\textrelation{\#}\neq\emptyset\}\cup\{c|e\in\textrelation{\#}, (e,c)\in\mathcal{K}_c\}$\\
     \textrelation{ARGMAX/ARGMIN} & $\{r|h\in\textrelation{\#}, t\in\mathcal{L}, (h, r, t)\in\mathcal{K}_r\}$\\
     \textrelation{LT(LE/GT/GE)} & $\{r|t<(\leq/ >/\geq)\textrelation{\#}, (h, r, t)\in\mathcal{K}_r\}$\\
     \textrelation{COUNT} & \{)\}\\
     \textrelation{CONS} & $\{(r, t)|h\in\textrelation{\#}, (h, r, t)\in\mathcal{K}_r\}$\\
     \textrelation{TC} & $\{(r, t)|h\in\textrelation{\#}, (h, r, t)\in\mathcal{K}_r, t\in\mathcal{L}\text{ is a temporal expression}\}$\\
    \bottomrule    
    \end{tabular}}
    \caption{A set of rules to expand a preceding subprogram given a function. The execution of the subprogram is denoted as \textrelation{\#}. These expansion rules reduce the search space significantly with a faithfulness guarantee. \textrelation{COUNT} takes no other argument, so the only admissible token is ``)".}
    \label{tab:cons}
    % \vspace{-10pt}
\end{table*}

\subsection{Dynamic Contextualized Encoding}
\label{sec:plm}
In semantic parsing, PLMs have typically been used to jointly encode the input question and all schema items via concatenation~\cite{hwang2019comprehensive, zhang2019editing, wang-etal-2020-rat}. However, direct concatenation is not feasible for KBQA due to a large number of schema items. Instead of obtaining a static representation for the question and items from $\mathcal{V}$ before decoding~\cite{gu2020iid,chen2021retrack}, we propose to do dynamic contextualized encoding at each decoding step; for each step, we use BERT to jointly encode the question and only the admissible tokens from $\mathcal{V}$. \OurMethod's dynamic program induction vastly reduces the number of candidate tokens at each step and allows us to concatenate the question and the admissible tokens into a compact sequence:\footnote{We omit the wordpieces tokenization here for brevity.}\\\\
\centerline{[CLS], $x_1$, ..., $x_{|q|}$, [SEP], $s^t_1$, ..., $s^t_m$, [SEP]}\\\\
where $\{s^t_i\}\subset \mathcal{V}$ are admissible tokens at step $t$ and $|\{s^t_i\}| = m$. After feeding the concatenated sequence to BERT, we obtain the question representation $\mathbf{Q_t} = (\mathbf{x_1}, ..., \mathbf{x_q})$ by further feeding the outputs from BERT to an LSTM encoder. For each admissible token, we represent it by averaging BERT outputs corresponding to its wordpieces. In this way, we also obtain the embedding matrix $\mathbf{W_t} \in \mathbb{R}^{m\times d}$, where each row corresponds to the embedding of an admissible token. The contextualized representation $\mathbf{Q_t}$ and $\mathbf{W_t}$ are both dynamically computed at each time step. Words and corresponding schema items are implicitly linked to each other via BERT's self-attention.

\subsection{Decoding}
We use an LSTM decoder. At decoding step $t$, given the hidden state $\mathbf{h_{t-1}}$ and input $\mathbf{c_{t-1}}$, we obtain the updated hidden state $\mathbf{h_t}$ by:
\begin{equation}
  \mathbf{h_t} = \textrm{LSTM}_\theta(\mathbf{h_{t-1}}, \mathbf{c_{t-1}})
\end{equation}
where our LSTM decoder is parameterized by $\theta$.

With $\mathbf{h_t}$ and $\mathbf{W_t}$---the embedding matrix of admissible tokens (determined by dynamic program induction)---we obtain the probability of generating a token from the admissible tokens:
\begin{equation}
    p(y_t=s_{ti}|q, y_{<t}) = [\textrm{Softmax}(\mathbf{W_t}\mathbf{h_t})]_i
\end{equation}

The input $\mathbf{c_{t}}$ for the next step is obtained via the concatenation of the contextualized embedding of the current output token and the weighted representation of the question based on attention:
\begin{gather}
    \mathbf{a_t} = \textrm{softmax}(\mathbf{Q_t}\mathbf{h_t}) \\
    \mathbf{q_t} = (\mathbf{a_t})^T\mathbf{Q_t} \\
    \mathbf{c_t} = [[\mathbf{W_t}]_j;\mathbf{q_t}]
\end{gather}
where $;$ denotes concatenation, and $j$ denotes the index of the predicted $y_t$ in $\mathbf{W_t}$.

\subsection{Training and Inference}
\label{sec:train}
% Different from~\citet{liang-etal-2017-neural}, 
We train \OurMethod with question-program pairs using cross entropy loss. The model learns to maximize the probability of predicting the gold token out of a small set of admissible tokens at each step, which is different from training a conventional Seq2Seq model using a static vocabulary.\nop{\footnote{We discuss different training schemes in Appendix~\ref{sec:scheme}.}}
% Execution-guided search space pruning does help to significantly reduce the number of candidate schema items during each decoding step, however, it also poses another challenge for training the model, i.e., the model tends to learn some shortcut instead of really interpret the question correctly using only the admissible actions for training (todo: elaborate). One straightforward way is to train the model with the entire vocabulary $\mathcal{V}$ at each decoding step and only do constrained decoding during inference, however, training the model in this way is prohibitively expensive, if feasible at all. To force the model to better understand the question and make it computationally effective at the same time, we propose a noise-adding training strategy. The idea is similar to adversarial training~\cite{}, i.e., during training we extend the set of admissible tokens with some schema items that are most similar to the gold token. 

During inference, \OurMethod assumes an entity linker to identify a set of entities from the question at the beginning of program induction. However, the entity linker may identify false entities. To deal with it, \OurMethod initiate its decoding process with different hypotheses from the set of entities. Basically, it tries out all possible combinations of the identified entities (i.e., the power set of the identified entities), considering that our entity linker normally can only identify no more than two entities from a question.

\section{Experimental Setup}

\begin{remark}[Datasets.]
We evaluate \OurMethod on three KBQA datasets covering diverse KB queries.\\
\noindent \textbf{\GrailQ}~\cite{gu2020iid} is a large-scale KBQA dataset that contains complex questions with various functions, including comparatives, superlatives, and counting. It evaluates the generalizability of KBQA at three levels: \iid, compositional and zero-shot. \nop{We thus use \GrailQ as our major evaluation dataset to validate the generalizability of \OurMethod. Its test set is hidden behind a leaderboard but the validation set is public. We will report the overall results on the test set but the fine-grained analyses will be based on the validation set.} \\
\noindent \textbf{\GraphQ}~\cite{su-etal-2016-generating} also contains questions of diverse nature. It is particularly challenging because it exclusively focuses on non-i.i.d. generalization.\footnote{\GraphQ originally uses \Freebase (version 2013-07) as their KB, while \GrailQ and \WebQ use \Freebase (version 2015-08-09). In ~\citet{gu2020iid}, programs in \GraphQ are converted into the corresponding \Freebase 2015-08-09 version, and we will use this version in our experiments.}\\
\noindent \textbf{\WebQSP}\cite{yih-etal-2016-value} is a clean subset of \WebQ\cite{berant-etal-2013-semantic} with annotated logical forms. All questions in it are collected from Google query logs, featuring more realistic and complicated information needs such as questions with temporal constraints.

The total number of questions in \GrailQ, \GraphQ, and \WebQ is 64,331, 5,166, and 4,737 respectively.\\
\end{remark}

\noindent \textbf{Evaluation Metrics.} For \GrailQ, we use their official evaluation script with two metrics, EM, i.e., program exact match accuracy, and F1, which is computed based on the predicted and the gold answer set. For \GraphQ and \WebQSP, we follow the standard practice and report F1.\\

\noindent \textbf{Models for Comparison.}
We compare \OurMethod with the previous best-performing models on three different datasets. For \GrailQ and \WebQSP, the state-of-the-art model is \textbf{RnG-KBQA}~\cite{ye2021rng}. Though RnG-KBQA uses T5 to decode the target program as unconstrained sequence transduction, it still heavily depends on candidate enumeration as a prerequisite. Therefore, it is not a generation-based model like ours. \textbf{ReTraCk}~\cite{chen2021retrack} is the state-of-the-art generation-based model on \GrailQ which poses grammar-level constraints to the decoder to generate well-formed but unnecessarily faithful programs. For \GraphQ, the ranking-based model \textbf{SPARQA}~\cite{sun2020sparqa} has achieved the best results so far. It uses BERT as a feature extractor for downstream classifiers. In addition to the state-of-the-art models, we also compare \OurMethod with \textbf{BERT+Transduction} and \textbf{BERT+Ranking}~\cite{gu2020iid}, which are two baseline models on \GrailQ that enhance a vanilla Seq2Seq model with BERT to perform generation and ranking respectively.\\

\nop{For \GrailQ, we compare \OurMethod with the strong baseline models proposed in the original paper~\cite{gu2020iid}, i.e., BERT+Transduction and BERT+Ranking. BERT+Transduction is a Seq2Seq model with BERT for joint encoding of the question and schema items, while BERT+Ranking uses the same Seq2Seq model but to rank candidate programs enumerated from the KB instead of generating a program. BERT+Ranking model is also the best-performing model on \GraphQ. The performance of QGG~\cite{lan-jiang-2020-query} on \GrailQ is reported as well. QGG is the best-performing model on several KBQA datasets including \WebQ. It uses beam search to prune the search space and uses BERT to provide features for ranking.

On \WebQ, we compare \OurMethod with the prior art QGG and several strong KBQA models proposed in recent years. On \GraphQ, in addition to BERT+Ranking, we also compare with SPARQA~\cite{sun2020sparqa} and several strong baselines reported in their paper. Similar to QGG, SPARQA also uses BERT to provide features to rank the candidate programs.\\}
% Note that, the most relevant model NSM is only implemented on \WebQ. 
\begin{table*}[th]
\small
\begin{subtable}{\linewidth}
\centering
\resizebox{\textwidth}{!}{\begin{tabular*}{\textwidth}{l@{\extracolsep{\fill}}cccccccc}
\toprule
& \multicolumn{2}{c}{\textbf{Overall}}   &\multicolumn{2}{c}{\textbf{\IID}}  &\multicolumn{2}{c}{\textbf{Compositional}}    &\multicolumn{2}{c}{\textbf{Zero-shot}}            \\ \cmidrule{2-9}
%  & \multicolumn{2}{c}{\textbf{Perfect Entity Linking}} & \multicolumn{2}{c}{\textbf{BERT Entity Linker}}               \\ \hline
% & \textbf{Exact Match} & \textbf{F1} & \\ 
\multicolumn{1}{l}{\textbf{Model}} & \textbf{EM} & \textbf{F1} & \textbf{EM} & \textbf{F1} & \textbf{EM} & \textbf{F1} & \textbf{EM} & \textbf{F1} \\ 
\midrule
\multicolumn{1}{l}{QGG$^*$~\cite{lan-jiang-2020-query}} & $-$    & 36.7  & $-$ & 40.5  & $-$   & 33.0  & $-$ &  36.6   \\
\multicolumn{1}{l}{BERT+Transduction$^*$~\cite{gu2020iid}} & 33.3  & 36.8  & 51.8  & 53.9      & 31.0  & 36.0      & 25.7  & 29.3    \\
\multicolumn{1}{l}{BERT+Ranking$^*$~\cite{gu2020iid}}      & 50.6  &  58.0     & 59.9  & 67.0       & 45.5  &   53.9    & 48.6  &  55.7  \\
\multicolumn{1}{l}{ReTraCk~\cite{chen2021retrack}}      & 58.1  &  65.3     & 84.4  & 87.5       & 61.5  &   70.9    & 44.6  &  52.5  \\
\multicolumn{1}{l}{RnG-KBQA$^*$~\cite{ye2021rng}}      & 61.4 & 67.4 & 78.0 & 81.8  & 55.0 & 63.2 & 56.7  & 63.0  \\
\multicolumn{1}{l}{ArcaneQA$^*$} & 58.8 & 67.2 & 77.8 & 81.6  & 58.0 & 66.1 & 50.4   & 61.8 \\
%\multicolumn{1}{l}{GloVe+Ranking~\cite{gu2020iid}}               & 39.5  & 45.1  & 62.2  & 67.3  & 40.0  & 47.8  & 28.9  & 33.8  \\
%\multicolumn{1}{l}{GloVe+Transduction~\cite{gu2020iid}}               & 17.6  & 18.4  & 50.5  & 51.6  & 16.4  & 18.5  & 3.0   & 3.1   \\ 

\midrule

\multicolumn{1}{l}{RnG-KBQA~\cite{ye2021rng}}      & \textbf{68.8}  &  \textbf{74.4}     & \textbf{86.2}  & \textbf{89.0}       & 63.8  &   71.2    & \textbf{63.0}  &  \textbf{69.2}  \\
% \multicolumn{1}{l}{\OurMethod} &\textbf{57.9} &\textbf{64.9} & \textbf{76.5} & \textbf{79.5} & \textbf{56.4} & \textbf{63.5} & \textbf{50.0} &\textbf{58.8}\\
\multicolumn{1}{l}{\OurMethod} &63.8 &73.7 & 85.6 & 88.9 & \textbf{65.8} & \textbf{75.3} & 52.9 &66.0\\
\multicolumn{1}{l}{\hspace{10pt} w/o contextualized encoding} & 49.7&  59.1&  77.6& 82.1 & 50.5 & 59.4 & 36.5 & 48.5\\

% \multicolumn{1}{r|}{$-$ BERT}               &   &   &   &  &  &   &   &  \\
\bottomrule
\end{tabular*}}
\caption{\GrailQ}
\label{table:overall_grail}
\end{subtable}
% \vspace*{0.5em}
% \newline
% \vspace*{1 cm}
% \newline
\begin{subtable}{0.525\linewidth}
\centering
 \resizebox{0.9\textwidth}{!}{\begin{tabular}{lc}
    \toprule
      \multicolumn{1}{l}{\textbf{Model}}  & \textbf{F1} \\
    \midrule
        UDEPLAMBDA \cite{reddy-etal-2017-universal} & 17.7 \\
        PARA4QA \cite{dong-etal-2017-learning-paraphrase} & 20.4 \\
        SPARQA \cite{sun2020sparqa} & 21.5 \\
        BERT+Ranking \cite{gu2020iid} & 25.0 (27.0) \\
    \midrule
        % \OurMethod & \textbf{28.1} (\textbf{30.3}) \\
        \OurMethod & \textbf{31.8} (\textbf{34.3}) \\
        \multicolumn{1}{l}{\hspace{10pt} w/o contextualized encoding} & 20.7 (22.4) \\
    \bottomrule
    \end{tabular}}
    \caption{\GraphQ}
\label{table:overall_gq}
\end{subtable}
\begin{subtable}{0.6\linewidth}
\centering
 \resizebox{.55\textwidth}{!}{\begin{tabular}{lc}
    \toprule
       \multicolumn{1}{l}{\textbf{Model}} & \textbf{F1} \\
    \midrule
        NSM \cite{liang-etal-2017-neural} & 69.0 \\
        KBQA-GST \cite{lan2019knowledge} & 67.9 \\
        TextRay \cite{bhutani2019learning} & 60.3 \\
        QGG \cite{lan-jiang-2020-query} & 74.0 \\
        ReTraCk \cite{chen2021retrack} & 71.0 \\
        CBR \cite{das2021case} & 72.8 \\
        RnG-KBQA \cite{ye2021rng} & \textbf{75.6} (74.5$^\sharp$)  \\
    \midrule
        \OurMethod & \textbf{75.6 (\textbf{75.6}$^\sharp$)} \\
        \multicolumn{1}{l}{\hspace{10pt} w/o contextualized encoding} & 68.8\\
    \bottomrule
    \end{tabular}}
        \caption{\WebQSP}
\label{table:overall_wq}
\end{subtable}
\caption{Overall results on three datasets. \OurMethod follows entity linking results from previous methods (i.e., RnG-KBQA's results on \GrailQ, QGG's results on \WebQSP, and~\citet{gu2020iid}'s results on \GraphQ) for fair comparison. Model names with $^*$ indicate using the baseline entity linking results on \GrailQ. $^\sharp$ In addition to using \WebQSP's official evaluation script, which sometimes considers multiple target parses for a question, we also report the performance when only the top-1 target parses are considered.}
\label{table:overall}
% \vspace{-15pt}
\end{table*}

% updated results no disamb: {'em': 0.5819665936059255, 'f1': 0.6725114215687972, 'em_iid': 0.7899185973700689, 'f1_iid': 0.8284107720833467, 'em_comp': 0.5726744186046512, 'f1_comp': 0.6662649248627602, 'em_zero': 0.4904192479469817, 'f1_zero': 0.6035583353072222}

% updated results with disabm: {'em': 0.6449247978232938, 'f1': 0.7331793555045433, 'em_iid': 0.8393863494051347, 'f1_iid': 0.875300736840114, 'em_comp': 0.6566537467700259, 'f1_comp': 0.7376735861631931, 'em_zero': 0.5502089036161937, 'f1_zero': 0.6657755476807448}

\noindent \textbf{Implementation.} Our models are implemented using PyTorch and AllenNLP~\cite{gardner-etal-2018-allennlp}. For BERT, we use the bert-base-uncased version provided by HuggingFace. For more details about implementation and hyper-parameters, we refer the reader to Appendix~\ref{sec:impl}.

\section{Results}
\subsection{Overall Evaluation}
We show the overall results in Table~\ref{table:overall} (for dev set results see \autoref{appendix:dev}). \OurMethod achieves the state-of-the-art performance on both \GraphQ and \WebQSP. For \GraphQ, there are 188 questions in \GraphQ's test set that cannot be converted into \Freebase 2015-08-09 version, so we treat the F1 of all those questions as 0 following~\citet{gu2020iid}, while the numbers in the parentheses are the actual F1 on the test set if we exclude those questions. \OurMethod significantly outperforms the prior art by over 10\%. The improvement over SPARQA shows the advantage of using PLMs for contextualized joint encoding instead of just providing features for ranking. On both \WebQSP and \GrailQ, \OurMethod also achieves the best performance or performs on par with the prior art in terms of F1. It outperforms ReTraCk by 4.3\% and 1.9\% (using the same entity linking results) on \WebQSP and \GrailQ respectively, suggesting that \OurMethod can more effectively reduce the search space with dynamic program induction compared with ReTraCk's grammar-based decoding. Also, our model performs on par with the previous state-of-the-art RnG-KBQA (i.e., same numbers on \WebQSP, while 0.7\% lower on \GrailQ). However, \OurMethod under-performs RnG-KBQA in EM on \GrailQ. The overall EM of \OurMethod is lower than RnG-KBQA by 5\%, and the gap on zero-shot generalization is even larger (i.e., around 10\%), despite the comparable numbers in F1. This can be explained by that \OurMethod learns to predict a program in a more flexible way and may potentially find some novel structures. \nop{For example, on \GrailQ's dev set, we find that for question \nl{supreme commander has influenced what?}, the gold program should be \textrelation{(AND cvg.computer\_videogame (JOIN cvg.computer\_videogame.influenced\_by m.07jxfk))}, while \OurMethod predicts \textrelation{(JOIN cvg.computer\_videogame.influenced\_inv (JOIN cvg.computer\_videogame.influenced\_by\_inv (JOIN cvg.computer\_videogame.influenced\_by m.07jxfk)))}, which can also retrieve the correct answer for the question. } This may further be supported by the observation that \OurMethod performs better than RnG-KBQA on compositional generalization, which requires KBQA models to generalize to unseen query structures during training. Overall, the results demonstrate \OurMethod's flexibility in handling KBQA scenarios of different natures.

\nop{
On \GrailQ, \OurMethod achieves a 6.9\% gain in overall F1 compared with the prior art. Moreover, \OurMethod outperforms all baseline models across all three levels of generalization. Notably, compared with BERT+Ranking, the F1 of \OurMethod is 12.5\% higher on \iid generalization, 9.6\% higher on compositional generalization, and 3.1\% higher on zero-shot generalization. BERT+Ranking is among the first to use PLMs in KBQA for strong generalizability, however, it relies on pre-defined templates to \textit{enumerate} candidate programs and thus can only handle programs of limited complexity and diversity. \OurMethod, through dynamic program induction, can easily handle much more complex and diverse questions (more on this in \S\ref{sec:fine-grained}). In addition, \OurMethod also exhibits a remarkable gain on the zero-shot KBQA dataset \GraphQ, which further confirms its strong generalizability. The improvement over SPARQA shows the advantage of using PLMs for contextualized joint encoding instead of just providing features for ranking. Note that, there are 188 questions in \GraphQ's test set that cannot be converted into \Freebase 2015-08-09 version, so we treat the F1 of all those questions as 0 following~\citet{gu2020iid}, while the numbers in the parentheses are the actual F1 on the test set if we exclude those questions. Last but not least, \OurMethod also performs competitively on \WebQ, even though this dataset mainly consists of simple questions (so dynamic program induction becomes less important) and evaluates \iid\ generalization (so dynamic contextualized encoding becomes less important). Overall, the results demonstrate that \OurMethod is a general model and can handle KBQA scenarios of different natures.
}

% results on webqsp:
% {'em': 0.649176327028676, 'f1': 0.724958755993077}

% \subsubsection{Fine-grained Evaluation}
% We conduct fine-grained analyses using the validation set of \GrailQ because its test set is hidden from the public. First, we break down the performance of \OurMethod based on the number of relations and the aggregation function of a question. As shown in Table, Then we also show \OurMethod's performance on different functions in Table
% \begin{figure}[t!]
%     \centering
%     \begin{subfigure}[t]{0.25\textwidth}
%         \centering
%         \includegraphics[height=1.2in]{acl-ijcnlp2021-templates/figs/att_hm.pdf}
%     \end{subfigure}%
%     ~ 
%     \begin{subfigure}[t]{0.25\textwidth}
%         \centering
%         \includegraphics[height=1.2in]{acl-ijcnlp2021-templates/figs/att_hm2.pdf}
%     \end{subfigure}
%     \caption{Implicit schema linking powered by dynamic contextualized encoding.}
%     \label{fig:att}
% \end{figure}
% \subsection{Key Designs for Generalizability}
\subsection{In-Depth Analyses}
\label{sec:fine-grained}
To gain more insights into \OurMethod's strong performance, we conduct in-depth analyses on the two key designs of \OurMethod.
\begin{remark}[Dynamic Program Induction.]
 One vanilla implementation of \OurMethod without dynamic program induction is BERT+Transduction, i.e., its search space and vocabulary during decoding is independent of previous predictions. As shown in Table~\ref{table:overall_grail}, when using the same entity linking results, \OurMethod outperforms BERT+Transduction by 30.4\% in overall F1 and is twice as good on zero-shot generalization. One major weakness of BERT+Transduction is that it predicts many programs that are not faithful to the KB, executing which will lead to empty answers. \nop{For example, for question \nl{What concert movie did Sundance Channel film?}, it predicts the program \textrelation{(AND film.film.music (JOIN Sundance\_Channel film.film.produced\_by)))}, 
 which evaluates to empty results on the KB because of the type mismatch of the arguments of \textrelation{AND}.} Note that post-hoc filtering by execution~\cite{wang2018robust} can only help to a limited degree due to the KB's broad schema, while this type of mistake is rooted out in \OurMethod by design.
 
%  which cannot be grounded to the KB because the execution of expression \textrelation{(JOIN Sundance\_Channel film.film.produced\_by)} should be films instead of music. This type of mistakes can be rooted out by \OurMethod.

Different from our search space pruning achieved with dynamic program induction, ranking-based models such as BERT+Ranking prunes unfaithful programs from their search space by ranking a set of faithful programs enumerated from the KB. These models typically make compromises on the complexity and diversity of programs during candidate enumeration. We break down the performance of \OurMethod on \GrailQ's validation set in terms of question complexity and function types and show the fine-grained results in Table~\ref{table:fine}. The comparison with BERT-Ranking demonstrates the scalability and flexibility of our dynamic program induction. We also compare with RnG-KBQA, which adopts exactly the same candidate enumeration module as BERT+Ranking, but it is enhanced with a T5-based revision module to edit the enumerated programs into more diverse ones. We observe that RnG-KBQA performs uniformly well across different programs except for programs with superlative functions (i.e., \textrelation{ARGMAX}/\textrelation{ARGMIN}), i.e., the F1 of it is lower than \OurMethod by over 50\%. This is because in their candidate generation step, there is no superlative function enumerated. Despite the effectiveness of their T5-based revision, their performance still heavily depends on the diversity of candidate enumeration, which restricts the flexibility of their method.
\begin{table*}[!h]
    \small
    \centering
    \resizebox{.6\textwidth}{!}{\begin{tabular}{lcccc}
    \toprule
    \textbf{Function} & \textbf{None} & \textbf{Count} & \textbf{Comparative} & \textbf{Superlative}\\\midrule
    BERT+Ranking & 59.1/66.0 & 43.0/53.2 & 0.0/14.5& 0/6.0\\
    RnG-KBQA &\textbf{77.5}/\textbf{81.8}& \textbf{73.0}/\textbf{77.5} &  \textbf{55.1}/\textbf{76.0} &13.8/22.3 \\
    \OurMethod &70.8/77.8 & 62.5/68.2& 54.5/75.7& \textbf{70.5}/\textbf{75.6} \\
     \midrule
     \textbf{\# of relations} & \textbf{1} & \textbf{2} & \textbf{3} & \textbf{4}\\\midrule
    BERT+Ranking  &57.4/61.5 & 39.8/54.7 & 0.0/22.9 & 0.0/25.0 \\
    RnG-KBQA &\textbf{75.7}/79.2 & \textbf{65.4}/\textbf{74.8}& \textbf{28.6}/\textbf{44.4} & \textbf{100.0}/\textbf{100.0} \\
    \OurMethod & 74.9/\textbf{80.9} &59.9/71.1& 27.6/37.7& \textbf{100.0}/\textbf{100.0}\\
    \bottomrule
    \end{tabular}}
\caption{Fine-grained results (EM/F1) on \GrailQ's dev set. \textbf{None} denotes programs with only \textrelation{AND} and \textrelation{JOIN}.}
\label{table:fine}
% \vspace{-10pt}
\end{table*}
\end{remark}

% \begin{table*}[t]
%     \centering
%     \small
%     \resizebox{\linewidth}{!}{\begin{tabular}{lc}
%     \toprule
%      \textbf{Example} & \textbf{Dataset} \\
%      \midrule
%      Question: & \multirow{2}{*}{\GrailQ}\\
%       Program: \textrelation{(AND food.cheese (AND (JOIN food.cheese.source\_of\_milk m.03fwl) (AND (JOIN food.cheese.source\_of\_milk m.01xq0k1) (JOIN food.cheese.texture m.02h87zr))))} & \\
%       \midrule
%       Question: & \multirow{2}{*}{\GrailQ}\\
%       Program: \textrelation{(AND food.cheese (AND (JOIN food.cheese.source\_of\_milk m.03fwl) (AND (JOIN food.cheese.source\_of\_milk m.01xq0k1) (JOIN food.cheese.texture m.02h87zr))))} & \\\midrule
    
%     \bottomrule    
%     \end{tabular}}
%     \caption{Examples of diverse programs generated by \OurMethod.}
%     \label{tab:mr}
%     \vspace{-10pt}
% \end{table*}

\begin{remark}[Dynamic Contextualized Encoding.]
To show the key role of dynamic contextualized encoding, we use GloVe~\cite{pennington-etal-2014-glove} to provide non-contextualized embeddings for both questions and tokens in $\mathcal{V}$. We fix GloVe embeddings during training to make the model less biased to the training distribution~\cite{gu2020iid} for \GrailQ and \GraphQ, which address non-\iid generalization, while for \WebQSP, we also update the word embeddings during training. Results in Table~\ref{table:overall_grail} show the importance of dynamic contextualized encoding, i.e., without contextualized encoding, the overall F1 decreases by 14.6\%, 11.1\%, and 6.5\% on three datasets respectively. We also notice that dynamic contextualized encoding is more critical for non-\iid generalization, i.e., on \GrailQ the F1 on \iid generalization only decreases by 6.8\%, while it decreases by 15.9\% and 17.5\% on compositional and zero-shot generalization. Without contextualized encoding, identifying the correct schema items from the KB in non-\iid setting is particularly challenging. Schema linking powered by dynamic contextualized encoding is the key to non-\iid\ generalization, which is a long-term goal of KBQA.

\nop{We also show the attention matrix at two different steps in predicting \textrelation{V2} and \textrelation{V3} in Figure~\ref{fig:intro} respectively (Figure~\ref{fig:att}), which clearly shows the link between the gold schema items and the question.}
% \begin{table}[H]
%     \small
%     \centering
%     \begin{tabular}{ccccc}
%     \toprule
%      & \textbf{Overall} & \textbf{I.I.D.} & \textbf{Compositional} & \textbf{Zero-shot}\\\midrule
%      \multicolumn{1}{l}{\OurMethod} &64.9 &79.5 &63.5 &58.8 \\
%      \multicolumn{1}{r}{w/o BERT} &45.6 &68.1 &46.7 &34.8\\
%     \bottomrule
%     \end{tabular}
% \caption{After removing the BERT module, the F1 of \OurMethod drops siginifcantly in non-\iid generalization on \GrailQ.}
% \label{table:glove}
% \end{table}
\end{remark}

% The above studies fully reveal that the performance of \OurMethod comes from the harmonious combination of the effective search space pruning strategy and the successful application of contexutal embeddings. Without the constrained-decoding strategy for search space pruning, the model suffers from scalability and flexibility, and without contexutal embeddings, the model performs poorly in non-\iid generalization. 

\subsection{Efficiency Analysis}
We compare the running time of \OurMethod and ranking-based models in the online mode (i.e., no offline caching) to mimic the real application scenario. To make the comparison fair, we configure all models to interact with the KB via the same Virtuoso SPARQL endpoint. We run each model on 1,000 randomly sampled questions and report the average running time per question on a GTX 2080 Ti card. As shown below, our model is faster than BERT+Ranking and RnG-KBQA by an order of magnitude, because \OurMethod dynamically prunes the search space and does not run the time-consuming queries for enumerating two-hop candidates.

\nop{The running time of QGG is comparable to \OurMethod partly because it uses beam search to reduce the search space, and partly because of its restrictions on the shape of the admissible logical forms (a core relational path plus constraints). Nonetheless, \OurMethod is still the most efficient.}

% \begin{table}[!h]
%     \small
%     \centering
%     \resizebox{0.5\textwidth}{!}{
%     \begin{tabular}{ccccccc}
%     \toprule
%          & BERT+Ranking & BERT+Transduction & QGG & \OurMethod \\\hline
%          \textbf{Running time (s)} & $115.5\pm107.9$& $60.9 \pm 47.8$ & $8.3 \pm 5.2$& $5.6\pm 2.8$ \\
%          \bottomrule
%     \end{tabular}}
%     \caption{Running time of different baselines.}
%     \label{table:time}
% \end{table}

\begin{table}[!h]
    \small
    \centering
    \resizebox{0.47\textwidth}{!}{
    \begin{tabular}{cccc}
    \toprule
         & \textbf{BERT+Ranking} & \textbf{RnG-KBQA}  & \textbf{\OurMethod} \\
         \midrule
         \textbf{Time (s)} & 115.5 & 82.1  & 5.6 \\
         \bottomrule
    \end{tabular}}
    \label{table:time}
\end{table}

% The results in Table~\ref{table:train} show that adding more noise during training does not lead to better performance. 

% \begin{table}[H]
%     \small
%     \centering
%     \begin{tabular}{cccc}
%     \toprule
%         & \textbf{EM} &\textbf{F1}\\
%     \midrule
%     Train w/ more noise & 55.4 & 63.6\\
%     \bottomrule
%     \end{tabular}
% \caption{The performance on \GrailQ when training \OurMethod with more negative tokens.}
% \label{table:train}
% \end{table}

% \subsection{Case Study}

\nop{
\subsection{Error Analysis}
We manually analyze 100 error cases randomly sampled from the validation set of \GrailQ to discuss venues for future improvement. We identify two main sources of errors:

\begin{remark}[Entity Errors (50\%):] Entity linking errors account for about half of the errors. For example, the entity linker incorrectly links \nl{John Elliott} in question \nl{How many game expansions has John Elliott released?} to the politician John Elliott instead of the game publisher. We currently adopt a pipelined design and use off-the-shelf entity linkers, and the correct entity is eliminated at an early stage by the entity linker. Integrating entity linking as an organic part of \OurMethod could be a promising way to improve both entity linking and semantic parsing, e.g., it is likely for \OurMethod to tell that the game publisher John Elliott is more plausible than the politician by looking at the context of the two entities during dynamic program induction.
\end{remark}

\begin{remark}[Schema Linking Errors (40\%):]
Even though dynamic contextualized encoding helps schema linking, it could still fail to identify the correct schema items for many questions, especially in zero-shot generalization with unseen schema items. More explicit schema linking could be a promising direction for future investigations.
\end{remark}
}

% \begin{remark}[Missing Constraint (10\%)]
% \GrailQ involves complex questions with multiple constraints, and sometimes some constraint might be missed by \OurMethod. For example, 
% \end{remark}

% \subsection{Generalization to a Different Knowledge Base}
% To further demonstrate the strong generalizability of our proposed method, we set up an interesting yet extremely challenging task, i.e., training on Freebase while testing on DBpedia. 
\section{Conclusions}
We present a novel generation-based KBQA model, \OurMethod, which simultaneously addresses the large search space and schema linking challenges in KBQA with dynamic program induction and dynamic contextualized encoding. Experimental results on several popular datasets demonstrate the advantages of \OurMethod in both effectiveness and efficiency. In the future, we will focus on developing generation-based KBQA models with stronger zero-shot generalizability. In addition, exploring other pre-trained language models such as T5~\cite{raffel2019exploring} for generation-based KBQA is also an interesting direction.

\section*{Acknowledgement}

The authors would like to thank the colleagues from the OSU NLP group and the anonymous reviewers for their thoughtful comments. This research was supported by NSF OAC 2112606.

\bibliographystyle{acl_natbib}
\bibliography{anthology,acl2021}

\begin{thebibliography}{43}
\expandafter\ifx\csname natexlab\endcsname\relax\def\natexlab#1{#1}\fi

\bibitem[{Abujabal et~al.(2017)Abujabal, Yahya, Riedewald, and
  Weikum}]{Abujabal}
Abdalghani Abujabal, Mohamed Yahya, Mirek Riedewald, and Gerhard Weikum. 2017.
\newblock \href {https://doi.org/10.1145/3038912.3052583} {Automated template
  generation for question answering over knowledge graphs}.
\newblock In \emph{Proceedings of the 26th International Conference on World
  Wide Web}, WWW ’17, page 1191–1200, Republic and Canton of Geneva, CHE.
  International World Wide Web Conferences Steering Committee.

\bibitem[{Bahdanau et~al.(2015)Bahdanau, Cho, and
  Bengio}]{DBLP:journals/corr/BahdanauCB14}
Dzmitry Bahdanau, Kyunghyun Cho, and Yoshua Bengio. 2015.
\newblock \href {http://arxiv.org/abs/1409.0473} {Neural machine translation by
  jointly learning to align and translate}.
\newblock In \emph{3rd International Conference on Learning Representations,
  {ICLR} 2015, San Diego, CA, USA, May 7-9, 2015, Conference Track
  Proceedings}.

\bibitem[{Berant et~al.(2013)Berant, Chou, Frostig, and
  Liang}]{berant-etal-2013-semantic}
Jonathan Berant, Andrew Chou, Roy Frostig, and Percy Liang. 2013.
\newblock \href {https://www.aclweb.org/anthology/D13-1160} {Semantic parsing
  on {F}reebase from question-answer pairs}.
\newblock In \emph{Proceedings of the 2013 Conference on Empirical Methods in
  Natural Language Processing}, pages 1533--1544, Seattle, Washington, USA.
  Association for Computational Linguistics.

\bibitem[{Berant and Liang(2014)}]{berant2014semantic}
Jonathan Berant and Percy Liang. 2014.
\newblock Semantic parsing via paraphrasing.
\newblock In \emph{Proceedings of the 52nd Annual Meeting of the Association
  for Computational Linguistics (Volume 1: Long Papers)}, pages 1415--1425.

\bibitem[{Bhutani et~al.(2019)Bhutani, Zheng, and
  Jagadish}]{bhutani2019learning}
Nikita Bhutani, Xinyi Zheng, and HV~Jagadish. 2019.
\newblock Learning to answer complex questions over knowledge bases with query
  composition.
\newblock In \emph{Proceedings of the 28th ACM International Conference on
  Information and Knowledge Management}, pages 739--748.

\bibitem[{Bhutani et~al.(2020)Bhutani, Zheng, Qian, Li, and
  Jagadish}]{bhutani-etal-2020-answering}
Nikita Bhutani, Xinyi Zheng, Kun Qian, Yunyao Li, and H.~Jagadish. 2020.
\newblock \href {https://doi.org/10.18653/v1/2020.nli-1.1} {Answering complex
  questions by combining information from curated and extracted knowledge
  bases}.
\newblock In \emph{Proceedings of the First Workshop on Natural Language
  Interfaces}, pages 1--10, Online. Association for Computational Linguistics.

\bibitem[{Bollacker et~al.(2008)Bollacker, Evans, Paritosh, Sturge, and
  Taylor}]{bollacker2008freebase}
Kurt Bollacker, Colin Evans, Praveen Paritosh, Tim Sturge, and Jamie Taylor.
  2008.
\newblock Freebase: a collaboratively created graph database for structuring
  human knowledge.
\newblock In \emph{Proceedings of the 2008 ACM SIGMOD international conference
  on Management of data}, pages 1247--1250.

\bibitem[{Cai and Yates(2013)}]{cai-yates-2013-semantic}
Qingqing Cai and Alexander Yates. 2013.
\newblock \href {https://www.aclweb.org/anthology/S13-1045} {Semantic parsing
  {F}reebase: Towards open-domain semantic parsing}.
\newblock In \emph{Second Joint Conference on Lexical and Computational
  Semantics (*{SEM}), Volume 1: Proceedings of the Main Conference and the
  Shared Task: Semantic Textual Similarity}, pages 328--338, Atlanta, Georgia,
  USA. Association for Computational Linguistics.

\bibitem[{Chen et~al.(2021)Chen, Liu, Yu, Lin, Lou, and
  Jiang}]{chen2021retrack}
Shuang Chen, Qian Liu, Zhiwei Yu, Chin-Yew Lin, Jian-Guang Lou, and Feng Jiang.
  2021.
\newblock Retrack: A flexible and efficient framework for knowledge base
  question answering.
\newblock In \emph{Proceedings of the 59th Annual Meeting of the Association
  for Computational Linguistics and the 11th International Joint Conference on
  Natural Language Processing: System Demonstrations}, pages 325--336.

\bibitem[{Chen et~al.(2019)Chen, Chang, Chen, Nayak, and Ku}]{chen2019uhop}
Zi-Yuan Chen, Chih-Hung Chang, Yi-Pei Chen, Jijnasa Nayak, and Lun-Wei Ku.
  2019.
\newblock Uhop: An unrestricted-hop relation extraction framework for
  knowledge-based question answering.
\newblock \emph{arXiv preprint arXiv:1904.01246}.

\bibitem[{Cheng et~al.(2019)Cheng, Reddy, Saraswat, and
  Lapata}]{cheng2019learning}
Jianpeng Cheng, Siva Reddy, Vijay Saraswat, and Mirella Lapata. 2019.
\newblock Learning an executable neural semantic parser.
\newblock \emph{Computational Linguistics}, 45(1):59--94.

\bibitem[{Das et~al.(2021)Das, Zaheer, Thai, Godbole, Perez, Lee, Tan,
  Polymenakos, and McCallum}]{das2021case}
Rajarshi Das, Manzil Zaheer, Dung Thai, Ameya Godbole, Ethan Perez, Jay-Yoon
  Lee, Lizhen Tan, Lazaros Polymenakos, and Andrew McCallum. 2021.
\newblock Case-based reasoning for natural language queries over knowledge
  bases.
\newblock \emph{arXiv preprint arXiv:2104.08762}.

\bibitem[{Devlin et~al.(2019)Devlin, Chang, Lee, and
  Toutanova}]{devlin-etal-2019-bert}
Jacob Devlin, Ming-Wei Chang, Kenton Lee, and Kristina Toutanova. 2019.
\newblock \href {https://doi.org/10.18653/v1/N19-1423} {{BERT}: Pre-training of
  deep bidirectional transformers for language understanding}.
\newblock In \emph{Proceedings of the 2019 Conference of the North {A}merican
  Chapter of the Association for Computational Linguistics: Human Language
  Technologies, Volume 1 (Long and Short Papers)}, pages 4171--4186,
  Minneapolis, Minnesota. Association for Computational Linguistics.

\bibitem[{Dong and Lapata(2016)}]{dong2016language}
Li~Dong and Mirella Lapata. 2016.
\newblock Language to logical form with neural attention.
\newblock In \emph{Proceedings of the 54th Annual Meeting of the Association
  for Computational Linguistics (Volume 1: Long Papers)}, pages 33--43.

\bibitem[{Dong et~al.(2017)Dong, Mallinson, Reddy, and
  Lapata}]{dong-etal-2017-learning-paraphrase}
Li~Dong, Jonathan Mallinson, Siva Reddy, and Mirella Lapata. 2017.
\newblock \href {https://doi.org/10.18653/v1/D17-1091} {Learning to paraphrase
  for question answering}.
\newblock In \emph{Proceedings of the 2017 Conference on Empirical Methods in
  Natural Language Processing}, pages 875--886, Copenhagen, Denmark.
  Association for Computational Linguistics.

\bibitem[{Gardner et~al.(2018)Gardner, Grus, Neumann, Tafjord, Dasigi, Liu,
  Peters, Schmitz, and Zettlemoyer}]{gardner-etal-2018-allennlp}
Matt Gardner, Joel Grus, Mark Neumann, Oyvind Tafjord, Pradeep Dasigi,
  Nelson~F. Liu, Matthew Peters, Michael Schmitz, and Luke Zettlemoyer. 2018.
\newblock \href {https://doi.org/10.18653/v1/W18-2501} {{A}llen{NLP}: A deep
  semantic natural language processing platform}.
\newblock In \emph{Proceedings of Workshop for {NLP} Open Source Software
  ({NLP}-{OSS})}, pages 1--6, Melbourne, Australia. Association for
  Computational Linguistics.

\bibitem[{Gu et~al.(2021)Gu, Kase, Vanni, Sadler, Liang, Yan, and
  Su}]{gu2020iid}
Yu~Gu, Sue Kase, Michelle Vanni, Brian Sadler, Percy Liang, Xifeng Yan, and
  Yu~Su. 2021.
\newblock Beyond iid: three levels of generalization for question answering on
  knowledge bases.
\newblock In \emph{Proceedings of the Web Conference 2021}, pages 3477--3488.

\bibitem[{Hwang et~al.(2019)Hwang, Yim, Park, and Seo}]{hwang2019comprehensive}
Wonseok Hwang, Jinyeung Yim, Seunghyun Park, and Minjoon Seo. 2019.
\newblock A comprehensive exploration on {WikiSQL} with table-aware word
  contextualization.
\newblock \emph{arXiv preprint arXiv:1902.01069}.

\bibitem[{Lan and Jiang(2020)}]{lan-jiang-2020-query}
Yunshi Lan and Jing Jiang. 2020.
\newblock \href {https://doi.org/10.18653/v1/2020.acl-main.91} {Query graph
  generation for answering multi-hop complex questions from knowledge bases}.
\newblock In \emph{Proceedings of the 58th Annual Meeting of the Association
  for Computational Linguistics}, pages 969--974, Online. Association for
  Computational Linguistics.

\bibitem[{Lan et~al.(2019{\natexlab{a}})Lan, Wang, and
  Jiang}]{lan2019knowledge}
Yunshi Lan, Shuohang Wang, and Jing Jiang. 2019{\natexlab{a}}.
\newblock Knowledge base question answering with topic units.(2019).
\newblock In \emph{Proceedings of the Twenty-Eighth International Joint
  Conference on Artificial Intelligence}, pages 5046--5052.

\bibitem[{Lan et~al.(2019{\natexlab{b}})Lan, Wang, and Jiang}]{lan2019multi}
Yunshi Lan, Shuohang Wang, and Jing Jiang. 2019{\natexlab{b}}.
\newblock Multi-hop knowledge base question answering with an iterative
  sequence matching model.
\newblock In \emph{2019 IEEE International Conference on Data Mining (ICDM)},
  pages 359--368. IEEE.

\bibitem[{Liang et~al.(2017)Liang, Berant, Le, Forbus, and
  Lao}]{liang-etal-2017-neural}
Chen Liang, Jonathan Berant, Quoc Le, Kenneth~D. Forbus, and Ni~Lao. 2017.
\newblock \href {https://doi.org/10.18653/v1/P17-1003} {Neural symbolic
  machines: Learning semantic parsers on {F}reebase with weak supervision}.
\newblock In \emph{Proceedings of the 55th Annual Meeting of the Association
  for Computational Linguistics (Volume 1: Long Papers)}, pages 23--33,
  Vancouver, Canada. Association for Computational Linguistics.

\bibitem[{Pennington et~al.(2014)Pennington, Socher, and
  Manning}]{pennington-etal-2014-glove}
Jeffrey Pennington, Richard Socher, and Christopher Manning. 2014.
\newblock \href {https://doi.org/10.3115/v1/D14-1162} {{G}lo{V}e: Global
  vectors for word representation}.
\newblock In \emph{Proceedings of the 2014 Conference on Empirical Methods in
  Natural Language Processing ({EMNLP})}, pages 1532--1543, Doha, Qatar.
  Association for Computational Linguistics.

\bibitem[{Raffel et~al.(2019)Raffel, Shazeer, Roberts, Lee, Narang, Matena,
  Zhou, Li, and Liu}]{raffel2019exploring}
Colin Raffel, Noam Shazeer, Adam Roberts, Katherine Lee, Sharan Narang, Michael
  Matena, Yanqi Zhou, Wei Li, and Peter~J Liu. 2019.
\newblock Exploring the limits of transfer learning with a unified text-to-text
  transformer.
\newblock \emph{arXiv preprint arXiv:1910.10683}.

\bibitem[{Reddy et~al.(2017)Reddy, T{\"a}ckstr{\"o}m, Petrov, Steedman, and
  Lapata}]{reddy-etal-2017-universal}
Siva Reddy, Oscar T{\"a}ckstr{\"o}m, Slav Petrov, Mark Steedman, and Mirella
  Lapata. 2017.
\newblock \href {https://doi.org/10.18653/v1/D17-1009} {Universal semantic
  parsing}.
\newblock In \emph{Proceedings of the 2017 Conference on Empirical Methods in
  Natural Language Processing}, pages 89--101, Copenhagen, Denmark. Association
  for Computational Linguistics.

\bibitem[{Rubin and Berant(2021)}]{rubin-berant-2021-smbop}
Ohad Rubin and Jonathan Berant. 2021.
\newblock \href {https://doi.org/10.18653/v1/2021.naacl-main.29} {{S}m{B}o{P}:
  Semi-autoregressive bottom-up semantic parsing}.
\newblock In \emph{Proceedings of the 2021 Conference of the North American
  Chapter of the Association for Computational Linguistics: Human Language
  Technologies}, pages 311--324, Online. Association for Computational
  Linguistics.

\bibitem[{Scholak et~al.(2021)Scholak, Schucher, and
  Bahdanau}]{scholak2021picard}
Torsten Scholak, Nathan Schucher, and Dzmitry Bahdanau. 2021.
\newblock Picard: Parsing incrementally for constrained auto-regressive
  decoding from language models.
\newblock \emph{arXiv preprint arXiv:2109.05093}.

\bibitem[{{Semantic Machines} et~al.(2020){Semantic Machines}, Andreas, Bufe,
  Burkett, Chen, Clausman, Crawford, Crim, DeLoach, Dorner, Eisner, Fang, Guo,
  Hall, Hayes, Hill, Ho, Iwaszuk, Jha, Klein, Krishnamurthy, Lanman, Liang,
  Lin, Lintsbakh, McGovern, Nisnevich, Pauls, Petters, Read, Roth, Roy, Rusak,
  Short, Slomin, Snyder, Striplin, Su, Tellman, Thomson, Vorobev, Witoszko,
  Wolfe, Wray, Zhang, and Zotov}]{SMDataflow2020}
{Semantic Machines}, Jacob Andreas, John Bufe, David Burkett, Charles Chen,
  Josh Clausman, Jean Crawford, Kate Crim, Jordan DeLoach, Leah Dorner, Jason
  Eisner, Hao Fang, Alan Guo, David Hall, Kristin Hayes, Kellie Hill, Diana Ho,
  Wendy Iwaszuk, Smriti Jha, Dan Klein, Jayant Krishnamurthy, Theo Lanman,
  Percy Liang, Christopher~H. Lin, Ilya Lintsbakh, Andy McGovern, Aleksandr
  Nisnevich, Adam Pauls, Dmitrij Petters, Brent Read, Dan Roth, Subhro Roy,
  Jesse Rusak, Beth Short, Div Slomin, Ben Snyder, Stephon Striplin, Yu~Su,
  Zachary Tellman, Sam Thomson, Andrei Vorobev, Izabela Witoszko, Jason Wolfe,
  Abby Wray, Yuchen Zhang, and Alexander Zotov. 2020.
\newblock \href {https://doi.org/10.1162/tacl_a_00333} {Task-oriented dialogue
  as dataflow synthesis}.
\newblock \emph{Transactions of the Association for Computational Linguistics},
  8:556--571.

\bibitem[{Su et~al.(2016)Su, Sun, Sadler, Srivatsa, G{\"u}r, Yan, and
  Yan}]{su-etal-2016-generating}
Yu~Su, Huan Sun, Brian Sadler, Mudhakar Srivatsa, Izzeddin G{\"u}r, Zenghui
  Yan, and Xifeng Yan. 2016.
\newblock \href {https://doi.org/10.18653/v1/D16-1054} {On generating
  characteristic-rich question sets for {QA} evaluation}.
\newblock In \emph{Proceedings of the 2016 Conference on Empirical Methods in
  Natural Language Processing}, pages 562--572, Austin, Texas. Association for
  Computational Linguistics.

\bibitem[{Sullivan(2020)}]{googlekg2020}
Danny Sullivan. 2020.
\newblock {A reintroduction to our Knowledge Graph and knowledge panels}.
\newblock
  \href{https://blog.google/products/search/about-knowledge-graph-and-knowledge-panels/}{blog.google}.

\bibitem[{Sun et~al.(2020)Sun, Zhang, Cheng, and Qu}]{sun2020sparqa}
Yawei Sun, Lingling Zhang, Gong Cheng, and Yuzhong Qu. 2020.
\newblock \href {https://arxiv.org/pdf/2003.13956.pdf} {Sparqa: Skeleton-based
  semantic parsing for complex questions over knowledge bases}.
\newblock In \emph{Proceedings of the Thirty-Fourth National Conference on
  Artificial Intelligence}, AAAI'20. AAAI Press.

\bibitem[{Sutskever et~al.(2014)Sutskever, Vinyals, and Le}]{Seq2Seq}
Ilya Sutskever, Oriol Vinyals, and Quoc~V Le. 2014.
\newblock \href
  {http://papers.nips.cc/paper/5346-sequence-to-sequence-learning-with-neural-networks.pdf}
  {Sequence to sequence learning with neural networks}.
\newblock In Z.~Ghahramani, M.~Welling, C.~Cortes, N.~D. Lawrence, and K.~Q.
  Weinberger, editors, \emph{Advances in Neural Information Processing Systems
  27}, pages 3104--3112. Curran Associates, Inc.

\bibitem[{Wang et~al.(2020)Wang, Shin, Liu, Polozov, and
  Richardson}]{wang-etal-2020-rat}
Bailin Wang, Richard Shin, Xiaodong Liu, Oleksandr Polozov, and Matthew
  Richardson. 2020.
\newblock \href {https://doi.org/10.18653/v1/2020.acl-main.677} {{RAT-SQL}:
  Relation-aware schema encoding and linking for text-to-{SQL} parsers}.
\newblock In \emph{Proceedings of the 58th Annual Meeting of the Association
  for Computational Linguistics}, pages 7567--7578, Online. Association for
  Computational Linguistics.

\bibitem[{Wang et~al.(2018)Wang, Tatwawadi, Brockschmidt, Huang, Mao, Polozov,
  and Singh}]{wang2018robust}
Chenglong Wang, Kedar Tatwawadi, Marc Brockschmidt, Po-Sen Huang, Yi~Mao,
  Oleksandr Polozov, and Rishabh Singh. 2018.
\newblock Robust text-to-sql generation with execution-guided decoding.
\newblock \emph{arXiv preprint arXiv:1807.03100}.

\bibitem[{Xie et~al.(2022)Xie, Wu, Shi, Zhong, Scholak, Yasunaga, Wu, Zhong,
  Yin, Wang et~al.}]{xie2022unifiedskg}
Tianbao Xie, Chen~Henry Wu, Peng Shi, Ruiqi Zhong, Torsten Scholak, Michihiro
  Yasunaga, Chien-Sheng Wu, Ming Zhong, Pengcheng Yin, Sida~I Wang, et~al.
  2022.
\newblock Unifiedskg: Unifying and multi-tasking structured knowledge grounding
  with text-to-text language models.
\newblock \emph{arXiv preprint arXiv:2201.05966}.

\bibitem[{Ye et~al.(2021)Ye, Yavuz, Hashimoto, Zhou, and Xiong}]{ye2021rng}
Xi~Ye, Semih Yavuz, Kazuma Hashimoto, Yingbo Zhou, and Caiming Xiong. 2021.
\newblock Rng-kbqa: Generation augmented iterative ranking for knowledge base
  question answering.
\newblock \emph{arXiv preprint arXiv:2109.08678}.

\bibitem[{Yih et~al.(2015)Yih, Chang, He, and Gao}]{yih-etal-2015-semantic}
Wen-tau Yih, Ming-Wei Chang, Xiaodong He, and Jianfeng Gao. 2015.
\newblock \href {https://doi.org/10.3115/v1/P15-1128} {Semantic parsing via
  staged query graph generation: Question answering with knowledge base}.
\newblock In \emph{Proceedings of the 53rd Annual Meeting of the Association
  for Computational Linguistics and the 7th International Joint Conference on
  Natural Language Processing (Volume 1: Long Papers)}, pages 1321--1331,
  Beijing, China. Association for Computational Linguistics.

\bibitem[{Yih et~al.(2016)Yih, Richardson, Meek, Chang, and
  Suh}]{yih-etal-2016-value}
Wen-tau Yih, Matthew Richardson, Chris Meek, Ming-Wei Chang, and Jina Suh.
  2016.
\newblock \href {https://doi.org/10.18653/v1/P16-2033} {The value of semantic
  parse labeling for knowledge base question answering}.
\newblock In \emph{Proceedings of the 54th Annual Meeting of the Association
  for Computational Linguistics (Volume 2: Short Papers)}, pages 201--206,
  Berlin, Germany. Association for Computational Linguistics.

\bibitem[{Yu et~al.(2018)Yu, Zhang, Yang, Yasunaga, Wang, Li, Ma, Li, Yao,
  Roman, Zhang, and Radev}]{yu-etal-2018-spider}
Tao Yu, Rui Zhang, Kai Yang, Michihiro Yasunaga, Dongxu Wang, Zifan Li, James
  Ma, Irene Li, Qingning Yao, Shanelle Roman, Zilin Zhang, and Dragomir Radev.
  2018.
\newblock \href {https://doi.org/10.18653/v1/D18-1425} {{S}pider: A large-scale
  human-labeled dataset for complex and cross-domain semantic parsing and
  text-to-{SQL} task}.
\newblock In \emph{Proceedings of the 2018 Conference on Empirical Methods in
  Natural Language Processing}, pages 3911--3921, Brussels, Belgium.
  Association for Computational Linguistics.

\bibitem[{Zelle and Mooney(1996)}]{zelle1996learning}
John~M Zelle and Raymond~J Mooney. 1996.
\newblock Learning to parse database queries using inductive logic programming.
\newblock In \emph{Proceedings of the thirteenth national conference on
  Artificial intelligence}, pages 1050--1055.

\bibitem[{Zettlemoyer and Collins(2005)}]{zettlemoyer2005learning}
Luke~S Zettlemoyer and Michael Collins. 2005.
\newblock Learning to map sentences to logical form: structured classification
  with probabilistic categorial grammars.
\newblock In \emph{Proceedings of the Twenty-First Conference on Uncertainty in
  Artificial Intelligence}, pages 658--666.

\bibitem[{Zhang et~al.(2019)Zhang, Yu, Er, Shim, Xue, Lin, Shi, Xiong, Socher,
  and Radev}]{zhang2019editing}
Rui Zhang, Tao Yu, He~Yang Er, Sungrok Shim, Eric Xue, Xi~Victoria Lin, Tianze
  Shi, Caiming Xiong, Richard Socher, and Dragomir Radev. 2019.
\newblock Editing-based sql query generation for cross-domain context-dependent
  questions.
\newblock \emph{arXiv preprint arXiv:1909.00786}.

\bibitem[{Zhong et~al.(2017)Zhong, Xiong, and Socher}]{wikisql}
Victor Zhong, Caiming Xiong, and Richard Socher. 2017.
\newblock Seq2sql: Generating structured queries from natural language using
  reinforcement learning.
\newblock \emph{arXiv preprint arXiv:1709.00103}.

\end{thebibliography}
% \bibliography{acl2021}
\clearpage
\appendix
 \section{Meaning Representation}
 \label{appendix:mr}
 We provide a detailed description of our defined functions for S-expressions in \autoref{tab:mr}. We provide annotations in S-expressions for several KBQA datasets, including \WebQSP, \GraphQ, and \ComplexQ (which we did not use for experiments). All data files annotated by us can be found in our \href{https://github.com/dki-lab/ArcaneQA.}{Github Repo}.
 \begin{table*}[t]
    \centering
    % \small
    \resizebox{\linewidth}{!}{\begin{tabular}{ccl}
    \toprule
     \textbf{Function} & \textbf{Arguments}  & \textbf{Returns} \\
     \midrule
     \textbf{\textrelation{JOIN}} & a set of entities $u\subset(\mathcal{E}\cup\mathcal{L})$ and a relation $r\in\mathcal{R}$ & all entities connecting to any $e\in u$ via $r$\\
     \textbf{\textrelation{AND}} & two set of entities $u1\subset\mathcal{E}$ and $u2\subset\mathcal{E}$ & the intersection of two entities sets.\\
     \textbf{\textrelation{ARGMAX/ARGMIN}} & a set of entities $u\subset\mathcal{E}$ and a numerical relation $r\in\mathcal{R}$& a set of entities from $u$ with the maximum/minimum value for $r$\\
     \textbf{\textrelation{LT(LE/GT/GE)}} & a numerical value $u\subset\mathcal{L}$ and a numerical relation $r\in\mathcal{R}$ & all entities with a value $<(\leq/ >/\geq) u$ for relation $r$ \\
     \textbf{\textrelation{COUNT}} & a set of entities $u\subset\mathcal{E}$ & the number of entities in $u$\\
     \textbf{\textrelation{CONS}} & a set of entities a set of entities $u\subset\mathcal{E}$, a relation $r\in\mathcal{R}$, and a constraint $c\in(\mathcal{E}\cup\mathcal{L})$ & all $e\in u$ satisfying $(e, r, c)\in\mathcal{K}_r$\\
     \textbf{\textrelation{TC}} & a set of entities a set of entities $u\subset\mathcal{E}$, a relation $r\in\mathcal{R}$, and a temporal constraint $c\in\mathcal{L}$ &all $e\in u$ satisfying $(e, r, c)\in\mathcal{K}_r$ \\
    \bottomrule    
    \end{tabular}}
    \caption{Detailed descriptions of functions defined in our S-expressions. We extend the definitions in~\citet{gu2020iid} by introducing two new functions \textrelation{CONS} and \textrelation{TC}. Also, we remove the function $R$ and instead represent the inverse of a relation by adding a suffix ``\textrelation{\_inv}" to it. Note that, for arguments in \textrelation{AND} function, a class $c\in\mathcal{C}$ can also indicate a set of entities which fall into $c$.}
    \label{tab:mr}
    \vspace{-10pt}
\end{table*}

% \section{S-expression Annotation}
% \label{sec:anno}
% Both \GraphQ and \WebQ provide structured annotations for queries in their dataset. For \GraphQ, they use graph query as their meaning representation, which is the same as \GrailQ. As a result, we can convert graph query in \GraphQ into S-expression following the same way as how graph query is converted into S-expression in \GrailQ. For \WebQ, each question is originally annotated with a relation chain and optionally a set of constraints, the expressivity of which is a subset of graph query, so we can readily convert it into graph query and then use the same script to further convert graph query into S-expression. However, one unique feature in \WebQ is that it contains questions involving constraints specified with non-named entities that are not able to be identified from the question by an entity linker. For example, the query corresponds to question \nl{Who is Justin Bieber's brother?} involves two entities, namely, \textrelation{Justin\_Bieber} and \textrelation{Male}, where \textrelation{Male} is not a named entity explicitly stated in the question. Instead of using \textrelation{JOIN} to express constraints involving these entities, we define a new function \textrelation{CONS} to treat them differently. Similarly, we also define a new function \textrelation{TC} to handle time constraint in \WebQ. We will release our annotations with detailed specifications upon acceptance.

\section{Implementation Details}
\label{sec:impl}
\subsection{Entity Linking Results}
For \GrailQ, we use the entity linking results provided by~\citet{ye2021rng}; for \GraphQ, we use the entity linking results provided by~\citet{gu2020iid}; for \WebQSP, we follow the entity linking results provided by~\cite{lan-jiang-2020-query}. In addition, we find that answer types can be a strong clue for \GrailQ, so we also predict a set of \Freebase classes for \GrailQ as a special type of entity using a BERT-based classifier. All entity linking results can be found in our \href{https://github.com/dki-lab/ArcaneQA}{Github Repo}.
\subsection{Entity Anonymization}
After identifying a set of entities, we do entity anonymization for \WebQ, i.e., we replace the entity mention with the type of the corresponding entity. For example, mention ``\textit{Barack Obama}" will be replaced by ``\textit{US president}". However, the entity linker might identify some false positive mentions, and anonymizing these mentions would lead to some critical information loss. To address this problem, we identify a set of common false positive mentions that contain important information about the question in training data. Such words include ``government", ``zip", etc. For mentions include these words, we do not do anonymization.  Doing entity anonymization is a common practice on \WebQ, which can normally bring some gain of around 1 to 2 percent in F1, while for \GrailQ and \GraphQ, we did not observe any improvement, so we keep the original entity mentions for these two datasets. 
\subsection{Hyper Parameters}
For \OurMethod, we are only able to train our model with batch size 1 due to the memory consumption, so we choose a workaround to set the number of gradient accumulations to 16. We use Adam optimizer with an initial learning rate of 0.001 to update our own parameters in BERT-based models. For BERT's parameters, we fine-tune them with a learning rate of 2e-5.
For \OurMethod w/o BERT, we train it with batch size 32 and an initial learning rate of 0.001 using Adam optimizer.
For both models, the hidden sizes of both encoder and decoder are set to 768, and the dropout rate is set to 0.5.
All hyper-parameters are manually tuned according to the validation accuracy on the development set. specifically, we do manual hyper-parameter search from [1e-5, 2e-5, 3e-5], [8, 16, 32], [0.0, 0.2, 0.5] to tune the learning rate of fine-tuning BERT, steps of gradient accumulation and dropout rate respectively.

\subsection{Number of Model Parameters}
Total numbers of trainable parameters of \OurMethod and \OurMethod w/o BERT are 123,652,608 and 261,900 respectively. The reason that the trainable parameters of \OurMethod w/o BERT are so few is that we freeze the GloVe embeddings for non-\iid generalization. The number of parameters becomes 121,205,100 if we take the GloVe embeddings into account.
\begin{table*}[t]
\small
\centering
\begin{tabular*}{\textwidth}{l@{\extracolsep{\fill}}cccccccc}
\toprule
& \multicolumn{2}{c}{\textbf{Overall}}   &\multicolumn{2}{c}{\textbf{\IID}}  &\multicolumn{2}{c}{\textbf{Compositional}}    &\multicolumn{2}{c}{\textbf{Zero-shot}}            \\ \cmidrule{2-9}
%  & \multicolumn{2}{c}{\textbf{Perfect Entity Linking}} & \multicolumn{2}{c}{\textbf{BERT Entity Linker}}               \\ \hline
% & \textbf{Exact Match} & \textbf{F1} & \\ 
\multicolumn{1}{l}{\textbf{Model}} & \textbf{EM} & \textbf{F1} & \textbf{EM} & \textbf{F1} & \textbf{EM} & \textbf{F1} & \textbf{EM} & \textbf{F1} \\ 
\midrule
\multicolumn{1}{l}{BERT+Ranking~\cite{gu2020iid}}      & 51.0  &  58.4     & 58.6  & 66.1       & 40.9  &   48.1    & 51.8  &  59.2  \\
\multicolumn{1}{l}{RnG-KBQA~\cite{ye2021rng}}      & \textbf{71.4}  &  76.8  & \textbf{86.7}      & \textbf{89.0}  &   61.7   & 68.9  &  \textbf{68.8} & \textbf{74.7} \\
\multicolumn{1}{l}{\OurMethod} &69.5 &\textbf{76.9} & 86.1 & 89.2 & \textbf{65.5} & \textbf{73.9} & 64.0 &72.8\\

% \multicolumn{1}{r|}{$-$ BERT}               &   &   &   &  &  &   &   &  \\
\bottomrule
\end{tabular*}
\caption{The results on the validation set of \GrailQ. The overall trend is basically consistent with the test set.}
\label{table:dev}
\end{table*}
\subsection{Other Details}
We summarize some other details in our implementation that are critical to reproduction.

We identify the literals in \GrailQ and \GraphQ using hand-crafted regular expressions. There are two types of literals, i.e., date time and numerical value. Our regular expressions can identify around 98\% of all literals.

During dynamic program induction of \OurMethod, we follow the rules in Table~\ref{tab:cons} to run SPARQL queries to retrieve the admissible schema items. However, in some rare cases, the execution of a subprogram may contain a tremendous number of entities For example, the execution of \textrelation{(JOIN USA people.person.nationality)} contains over 500,000 entities, and running SPARQL queries for all entities in them is infeasible. As a result, we only run SPARQL queries for 100 entities sampled from the execution results. One better choice could be to use some more efficient indexing to query the KB instead of using SPARQL.

We construct the vocabulary $\mathcal{V}$ for different datasets in different ways. For \GrailQ, following~\citet{gu2020iid}, we construct the vocabulary using schema items from \Freebase \Commons. For \GraphQ, we construct the vocabulary using schema items from the entire \Freebase. For \WebQ, because it evaluates \iid generalization, so we construct the vocabulary from its training data. 

\section{Results on the Validation Set of \GrailQ}
\label{appendix:dev}
We show the results of \OurMethod, BERT+Ranking, and RnG-KBQA on the validation set of \GrailQ in Table~\ref{table:dev}. We observe that \OurMethod achieves a better F1 than RnG-KBQA. Overall, the trend is consistent with the test set. We also observe that the EM of \OurMethod on zero-shot generalization is significantly higher than the test set, which is interesting and remains for further investigation.

\nop{
\section{Training Scheme}
\label{sec:scheme}
% In our proposed model, we do the same constrained decoding for both training and inference, i.e., we train \OurMethod by maximizing the probability of predicting the target token from only a small set of admissible tokens at each step. 
\OurMethod is only trained with a small set of admissible tokens as negative examples, which might lead to under-training. We conduct experiments to compare our training scheme  with training using more negative examples. Specifically, we also train \OurMethod in a way similar to ~\citet{gu2020iid}, i.e., we extend the admissible tokens with the schema items that are reachable from the topic entity within two hops at each decoding step during training. The model trained in this way achieves an overall EM of 55.4 and F1 of 63.6 on the test set, which is slightly worse than the original model. One possible explanation is that adding more admissible tokens during training lead to a mismatch in the distribution of contextualized encoding between training and inference, so we also try to add the same negative tokens during the encoding phase of inference but mask them before the softmax layer, and we achieve 55.6 in EM and 63.7 in F1. These numbers reveal that our proposed training scheme is effective. However, the best practice for training a model with constrained decoding using strong supervision remains an open question, and we will leave it to future research.}
% \section{some more examples maybe?}

\end{document}